\documentclass{article}
\usepackage{microtype}
\usepackage{pdflscape} 
\usepackage{pdfpages} 

\usepackage{multirow}

\usepackage{graphicx}

\usepackage{booktabs} 
\usepackage{placeins} 
\usepackage{tabularx} 
\usepackage{svg}
\usepackage{longtable}
\usepackage{hyperref}

\usepackage[accepted]{icml2025}

\usepackage{amsmath}
\usepackage{amssymb}
\usepackage{mathtools}
\usepackage{amsthm}
\usepackage{subcaption}

\usepackage[capitalize,noabbrev]{cleveref}

\theoremstyle{plain}

\theoremstyle{definition}

\theoremstyle{remark}

\usepackage[textsize=tiny]{todonotes}
\usepackage{authblk} 

\title{Identifying Sparsely Active Circuits Through Local Loss Landscape Decomposition}

\author[1]{Brianna Chrisman\thanks{Email:brianna.chrisman@gmail.com}}
\author[2]{Lucius Bushnaq}
\author[2]{Lee Sharkey}

\affil[1]{Independent}
\affil[2]{Apollo Research}

\begin{document}

\maketitle



\begin{abstract}
Much of mechanistic interpretability has focused on understanding the activation spaces of large neural networks. However, activation space-based approaches reveal little about the underlying circuitry used to compute features. To better understand the circuits employed by models, we introduce a new decomposition method called \textbf{Local Loss Landscape Decomposition (L3D)}. L3D identifies a set of low-rank subnetworks—directions in parameter space—of which a subset can reconstruct the gradient of the loss between any sample's output and a reference output vector. We design a series of progressively more challenging toy models with well-defined subnetworks and show that L3D can nearly perfectly recover the associated subnetworks. Additionally, we investigate the extent to which perturbing the model in the direction of a given subnetwork affects only the relevant subset of samples. Finally, we apply L3D to a real-world transformer model and a convolutional neural network, demonstrating its potential to identify interpretable and relevant circuits in parameter space.

\end{abstract}

\section{Background}

Mechanistic interpretability aims to uncover the internal mechanisms responsible for the behavior of large models, enabling developers to better understand, intervene in, and align models \cite{bereska2024mechanistic}. One goal of the field is to decompose model behavior into subcomponents that are less complex and more human-interpretable while still fully explaining a model's behavior. The most popular method in this space is Sparse Dictionary Learning (SDL) \cite{cunningham2023sparse,bricken2023towards,gao2024scaling}, which identifies latent features by decomposing a model's activation space into an overcomplete basis of sparsely activating components. These learned basis vectors represent distinct features that can then be used to reconstruct the original activations.

\subsection{From Activation to Parameter-Based Interpretability}\label{subsec:activation_to_parameter}

However, decomposing the activation space of a model has various limitations. Current SDL algorithms struggle with reconstructing features of certain geometries, such as nonlinear features, feature manifolds, and certain types of superposition \cite{engels2024not,engels2024decomposing,merullo2024talking,lindsey2024sparse}. Such issues could become more pronounced in models with a less clearly defined read/write stream, such as diffusion models and recurrent networks. \cite{pascanu2013difficulty,ho2020denoising}. Additionally, activation space captures the \textit{features} extracted by a model's underlying circuits, but it says little about what mechanisms derived them.

Alternatively, to understand a model's underlying \textit{mechanisms}, we might interpret models through the lens of \textit{parameter space}. Parameters are the fundamental objects updated during training, and can capture information about a model's internal mechanisms, the training process, and the mechanistic relationship between outputs. We hypothesize that parameter space can hold interpretable units of computation \cite{sharkey2025open}: models can be decomposed into simpler \textit{subnetworks}, where each subnetwork is involved in the predictions of a subset of training data. To understand how we might go about identifying such sparsely active subnetworks, we first must understand some key insights about loss landscape geometry.

\subsection{Loss Landscape Geometry}\label{subsec:loss_landscape_geometry}

Singular learning theory (SLT) describes how the structure of parameter space influences model behavior \cite{watanabe2000algebraic,watanabe2005algebraic} and has been used to characterize model topologies \cite{bushnaq2024using,lau2023local} as well as different phases of the training process \cite{wang2024loss,hoogland2024developmental,davies2023unifying}. A key insight from SLT is that large models are highly degenerate in parameter space: they can have many different parameter configurations that achieve minimal loss on the training set \cite{wei2022deep,watanabe2007almost}. In fact, gradient descent tends to converge on configurations with many of these degenerate directions. Our work extends this hypothesis one step further:\textbf{ If models are highly degenerate with respect to the full training distribution, then with respect to a subset of the training data, they likely exhibit additional subset-specific degeneracies.}

Another key phenomenon our method relies on is that, at least in the current set of foundation models, local attribution methods appear to be good approximations of global relationships between pairs of samples. For example, attribution patching can successfully modify a model’s output by targeting specific activations determined by the first-order gradients of paired outputs \cite{nanda2023attribution,kramar2024atp,syed2023attribution}. Similarly, steering vectors—derived from differences in activations between paired samples—can effectively guide models toward specific behaviors, even when applied beyond the original magnitude of those activation differences \cite{turner2023,subramani2022extracting}. 

\subsection{Loss Landscape Decomposition}\label{subsec:loss_landscape_decomp}
Our goal in this paper is to identify directions in parameter space that correspond to subnetworks, as defined in Section \ref{subsec:activation_to_parameter}.

Two existing methods in particular address a similar problem of decomposing models into subnetworks. An earlier work \cite{matena2023npeff} decomposes parameter space by computing principal directions of a per-sample Fisher Information matrix to identify meaningful features. A more recent method, Attribution Parameter Decomposition \cite{braun2025interpretability}, decomposes model weights by identifying subnetworks where (1) the sum of subnetwork weights approximates the original model parameters, (2) for any given input, the outputs of the sum of topk-attributed networks has low behavioral loss when compared to those of the original model, and (3) subnetworks are individually simpler than the whole network. 

Rather than the parameter values themselves \cite{braun2025interpretability} or an approximate second order gradient of the parameters \cite{matena2023npeff}, \textbf{the object we decompose is the gradient of the loss between a sample's output and a reference output. }We aim to identify directions in parameter space that strongly affect this loss for some samples, and have little effect on the loss for other samples. 

\begin{figure}
  \begin{subfigure}{\columnwidth}
  \centering
  \includegraphics[width=.7\textwidth]{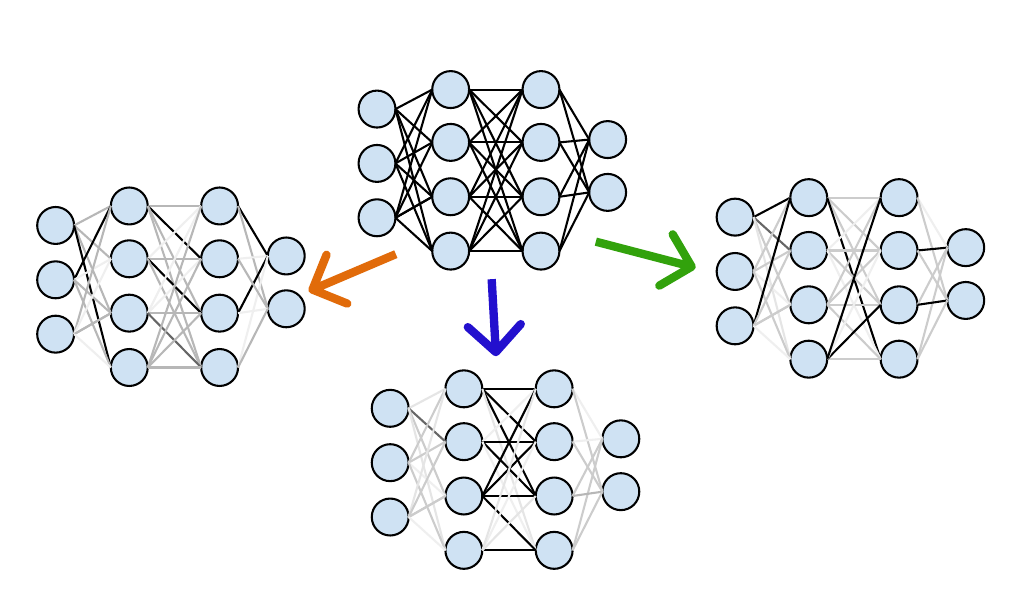}
  \end{subfigure}
  \begin{subfigure}{\columnwidth}
  \centering
  \includegraphics[width=\textwidth]{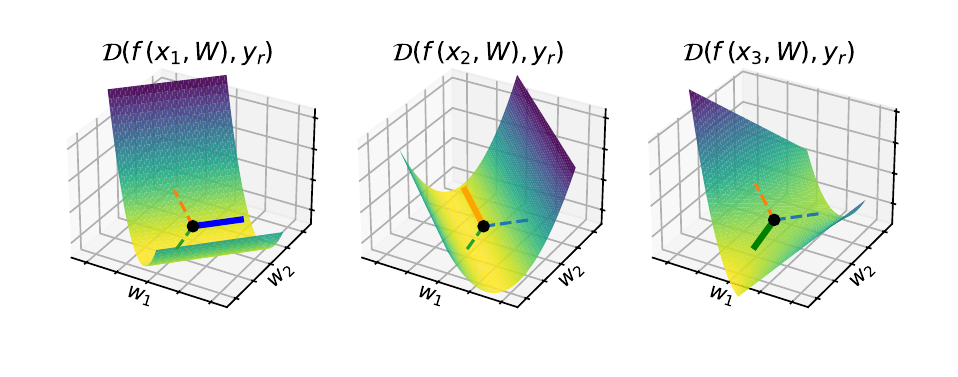}
  \end{subfigure} \caption{Decomposing a loss landscape into a set of parameter directions, or subnetworks, where a smaller subset of directions can approximately reconstruct the gradient of divergence/loss between any sample's output and a reference output. Here, $D$ is a loss, or \textit{divergence} measure, $f$ is our model, $W$ is the set of parameters in the model, $x_i$ is a sample input, and $y_r$ is a reference output}\label{fig:1_jacobian_diagram}
\end{figure}

In practice, we choose the reference output as another output sampled from the training distribution, and we discuss our reasoning in Section \ref{subsec:divergences}. Our goal is to identify low-rank directions in parameter space—henceforth referred to as subnetworks—such that for any pair of samples, a small number of these directions can be used to reconstruct this gradient (Figure \ref{fig:1_jacobian_diagram}).

We call our decomposition method Local Loss Landscape Decomposition (L3D). In this work, we first describe the mathematical foundation of our approach. We then develop progressively more complex toy models to evaluate the efficacy of L3D and characterize its limitations. Finally, we present preliminary results on real-world models to demonstrate L3D’s potential for scaling beyond toy settings.


\section{Methodology}\label{sec:methods}

In the next sections, we will formally set up our decomposition problem (Section \ref{subsec:setup}), define the criteria that we will use for our subnetwork/parameter directions (Section \ref{subsec:divergences}), describe how to efficiently decompose parameters into these directions (Section \ref{subsec:decomposition}), walk through our training algorithm (Section \ref{subsec:training}), and then explain how to use these decompositions to intervene on a model's behavior (Section \ref{subsec:intervention}). 

\textbf{From now on, we will use the word ``subnetworks" to refer to the directions in parameter space we wish to learn. 
}
\subsection{Set up}\label{subsec:setup}

Consider a model $f$ that takes a batch of inputs $X$ (with number of samples $n_s$ and input dimension $n_i$) and parameter values of $W$, and computes a batch of outputs (with output dimension $n_o$).
\begin{equation}
  f(x, W) : \mathbb{R}^{n_s \times n_i} \rightarrow \mathbb{R}^{n_s \times n_o}
\end{equation}
Our approach assumes that for a given input, there are many components of a model's parameters that are not involved in inference. Changing parameters in the direction of these components will not change the model's output. Conversely, changing parameters in the direction of components that \textit{are} involved \textit{would} change the model's output. Moreover, we are interested in finding parameter directions that, when perturbed, \textit{meaningfully} change a model's output. The next section will explain what constitutes ``meaningful."

\subsection{Divergences of Paired Outputs}\label{subsec:divergences}

Intervening in a relevant parameter direction should move a sample’s output either closer to or further from a \textbf{reference output}. This reference output should serve as a neutral and representative baseline that captures the typical behavior of the model’s output distribution. We considered three candidates for this reference:

\begin{enumerate}
  \item \textbf{A uniform output:} This reference consists of a vector with uniform values. However, it fails to account for the training distribution, leading to a bias toward learning subnetworks that influence outputs that skew toward particularly high or low values.
  \item \textbf{Mean of outputs:} This reference is computed by averaging each output index across the training distribution or a batch. While it is grounded in the data, it risks averaging away meaningful correlations between outputs, producing a reference that may still be out-of-distribution relative to the training data.
  \item \textbf{Another sample as the reference:} For each sample, we use the output of a randomly selected sample as the reference. This approach preserves the nuances of the output distribution but may lead to slow convergence due to high variance in reference selection.
\end{enumerate}

We thought (3) was the most principled, and least biased of the three. Although not tested rigorously, in early prototypes all three choices seemed to produce reasonable results on toy models and we did not find any issues with convergence using (3). For this work we use (3) as our reference output, but we believe other choices are possible and may have different strengths and weaknesses.

Therefore, we decompose gradients of the loss between pairs of outputs with the aim of finding directions that move a model's output towards or away from the reference. \textbf{Because we use the term ``loss" later on when we describe our training process, we will refer to this metric instead as ``divergence."}

The gradient of the divergence of a sample's output and a reference can be written as as:
\begin{align}\label{eq:divergence}
  &\nabla_W D(f(x_i, W), y_r)|_{W=W_0} \\
  &\text{ where } x_i \in X , y_r \in f(X)\notag
\end{align}
Here, $D$ is a divergence measure, $f$ is our model, $x_i$ is our input of interest, $y_r$ is a reference output (chosen as another output sampled from the training distribution), $W$ is a set of parameters and $W_0$ is the model's original parameters. Our toy models are regression-type models, so we use normalized MSE as divergence. For the real-world transformer and CNN models, which output probabilities, we used KL-divergence.

We abbreviate the expression in Eq. \ref{eq:divergence} as $\nabla_W D$.

\subsection{Sparse Principal Directions}\label{subsec:decomposition}

We want to decompose our per-sample gradients with respect to parameters into low-rank components. Each sample's gradient should be able to be expressed as a linear combination of a small set of these components. We will do this by learning transforms $V^{in} \in R^{n_v \times n_w}$ and $V^{out} \in R^{n_w \times n_v}$ where $n_w$ is the number of parameters in the model, and $n_v$ is the number of components (subnetworks) we wish to use to represent the parameter space. (For those familiar with the sparse dictionary learning set up, this is similar to learning a transform from activation space into feature space, and vice versa). 

$V^{in}$ effectively transforms a gradient from the parameter space to the subnetwork space, so that: 
\begin{equation}
  \nabla_V D = V^{in} \nabla_W D
\end{equation}
We want to find $V^{in}$ and $V^{out}$ such that for any given pair of samples, a small subset of subnetworks can approximately reconstruct the gradient of divergence. 
\begin{align}
  & \nabla_W D \approx V^{\text{out}} \Lambda V^{\text{in}} \nabla_W D \\
  & \text{where }
  \Lambda_{i,j} = \notag
  \begin{cases} 
    1 & \text{if } i = j \text{ and } i \in \underset{i}{\text{argTopK}} \left( \left| \nabla_{v_i} D \right| \right) \\
    0 & \text{otherwise}
  \end{cases} \notag
\end{align}

$\text{argTopk}$ relies on a hyperparameter $k$ that controls the number of components we wish to use to reconstruct each sample. In practice, we use a $\text{batchTopK}$ \cite{bussmann2024batchtopk} and a fraction for the $k$ hyperparameter rather than an absolute number. $k=0.1$ means that we select the top 10\% of $\nabla_V D$ magnitudes over $v$ and $x$ to reconstruct our batch of gradients.

\subsubsection{Low Rank parameter directions}
Learning a set of full rank parameter directions would be extremely expensive. We also expect that modular, sparsely active circuits would be lower rank than their full-model counterparts because they are processing smaller numbers of features. Therefore, we use low-rank representations of our $V^{in}$ and $V^{out}$, and correspondingly learn low-rank circuits (Appendix \ref{sec:low_rank}).  Specifically, we use a Tucker decomposition described in Section \ref{sec:low_rank}.

\subsection{Training}\label{subsec:training}
We wish to learn the decomposition-related transforms $V^{in}$ and $V^{out}$ that minimize the $\text{batchTopK}$ reconstruction loss of our divergence gradient described above. We use a (normalized) L2 norm loss.

\begin{equation}
  L = \frac{{|| \nabla_W D -  V^{out} \Lambda V^{in} \nabla_W D ||}_2}{{|| \nabla_W D ||}_2}
\end{equation}
For each batch of samples, we randomly select a reference sample $x_r$ to be paired with each sample $x_i$ in the batch. We then compute the gradient of divergence between $f(x_i)$ and $f(x_r)$ at the target model's parameters $W_0$. We transform that gradient into the subnetwork space using $V^{in}$, and compute the $\text{topK}$ components. We transform those components back into the original parameter space using $V^{out}$, and compute the loss between the reconstructed gradient and the original gradient. We apply a learning update to $V^{in}$ and $V^{out}$ with the goal of minimizing this loss. We also normalize $V^{out}$ to be a unit vector after each update in order to keep the magnitudes of $V^{in}$ and $V^{out}$ similar.

\begin{algorithm}\label{alg:training}
\caption{L3D algorithm for learning $V_{in}$ and $V_{out}$ transforms of parameter space.}
\begin{algorithmic}[1]
\FOR{each epoch}
  \FOR{each minibatch $X$}
  \FOR {each $x_i \in X$}
    \STATE Randomly select $x_r \in X$
    \STATE $\nabla_w D_i = \nabla_w D(f(x_i, W), f(x_r))|_{W=W_0}$
  \ENDFOR
  \STATE $\nabla_v D = {V^{in}} \nabla_w D$
  \STATE $\tau = \text{topK}(\text{abs}(\nabla_v D))$
  \STATE $\hat{\nabla}_w D = {V^{out}} (\nabla_v D \odot (\text{abs}(\nabla_v D) > \tau))$
  \STATE $L = \frac{{|| \nabla_w D - \hat{\nabla}_w D ||}_2}{{||\nabla_w D||}_2}$
  \STATE $L.\text{backward()}$
  \STATE Update $V^{in}$ and $V^{out}$
  \STATE Normalize $V^{out}$ to be unit vectors.
  \ENDFOR
\ENDFOR
\end{algorithmic}
\end{algorithm}

\subsection{Measuring and Intervening}\label{subsec:intervention}

Our learned subnetworks will just be the columns of $V^{out}$, restructured into the same tensor structure as $W$. After identifying subnetworks, we may want to intervene on a specific circuit.

If we wish to ``intervene" on a model using a single subnetwork, we can update the model's parameters by moving them in their unit direction, multiplied by a scalar factor ($\delta$). To tune our model in the direction of subnetwork $v_i$ and compute predictions on $x$, we evaluate:
\begin{equation}
  f(x, W + \delta v_i)
\end{equation}
We also may want to quantify the impact of a subnetwork in on a certain sample. First, we can compute the impact of a subnetwork on a specific output's ($f(x_i)$) divergence with a single reference output $y_j$. The impact $I$ of subnetwork $v_k$ on the gradient of divergence between $f(x_i)$ can be measured by:
\begin{equation}
  I(x_i, y_j, v_k) = \\
  \left| V^{in}_{k,:} \nabla_w D(f(x_i, W), y_j) \right|
\end{equation}

Because we are randomly sampling outputs from our training distribution as the reference output, we then average the impacts of a subnetwork $v_k$ and an input $x_i$ over many different reference samples to better quantify the impact of the subnetwork on a single sample's predictions overall. Although more computationally expensive, this gives a more robust measurement for the impact of a subnetwork on a specific sample. 
\begin{equation}
  I(x_i, v_k) = \frac{1}{n_j} \sum_{j=1}^{n_j} I(x_i, y_j, v_k)
\end{equation}
\section{Results}\label{sec:results}

To evaluate L3D's ability to decompose models, we focused on developing toy models that involve well-defined subnetworks.

We designed several toy models to test the efficacy of L3D. Our toy models all consist of several well-characterized computations being performed by the same model, with an sparse input space designed in a way that only a small number of computations are being performed for each input sample.  

Our toy models progressively test more complex types of circuits. Table \ref{tab:toy_models} describes our 4 toy models and the different attributes of circuitry the are designed to capture. The specific hyperparameters used to train our toy models are described in Appendix \ref{sec:toymodel_hyperparams}, as well as the hyperparameters used for each decomposition in Appendix \ref{sec:L3D_hyperparams}

\begin{table*}[htb]
  \centering
  \begin{tabularx}{\textwidth}{X X X X X} 
  \toprule
   & Toy model of superposition & Circuit Superposition (TMCS) & Higher Rank Circuit Superposition & Complex Loss Landscape \\ 
  \midrule

  \hline
  & \includegraphics[width=0.12\textwidth]{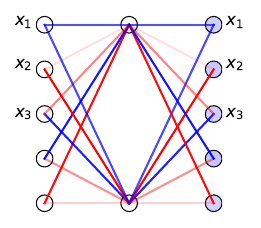} &
  \includegraphics[width=0.2\textwidth]{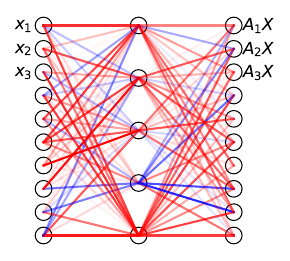} &
  \includegraphics[width=0.2\textwidth]{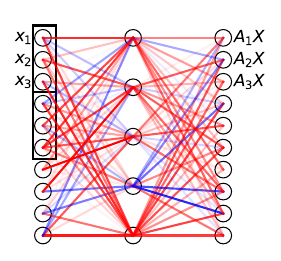} &
  \includegraphics[width=0.2\textwidth]{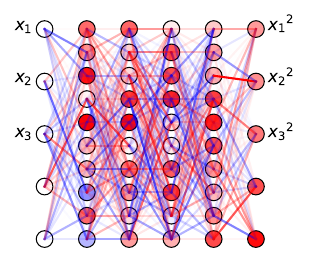} \\
   & $X \mapsto X$ & $X \mapsto A X$ & $X \mapsto A X$ & $X \mapsto X^2$ \\ 
  Feature Superposition & $\checkmark$ & $\checkmark$ & $\checkmark$ & $\checkmark$ \\ 
  Circuit Superposition & $\times$ & $\checkmark$ & $\checkmark$ & $\checkmark$ \\ 
  Circuits $>$ rank 1 & $\times$ & $\times$ & $\checkmark$ & probably $\checkmark$ \\ 
  Complex Loss Landscape & $\times$ & $\times$ & $\times$ & $\checkmark$ \\ 
  \bottomrule
  \end{tabularx}
  \caption{Our toy models and their various properties. Toy models are designed to test progressively more complicated phenomenon present in model circuitry,}
  \label{tab:toy_models}
\end{table*}

\subsection{Toy Model of Superposition}

\subsubsection{Setup}

We started off by validating our algorithm on a well-studied toy problem, the toy model of superposition (TMS). TMS is simple linear autoencoder with a low-dimensional hidden layer followed by a ReLU activation function at the output \cite{elhage2022toy}. The model is trained on a dataset of samples where few features are active at a time, and ``superimposes" these features in the hidden layer such that features' embeddings in the hidden layer have minimal interference with each other. We trained a toy model of superposition with 5 features and 2 hidden dimensions (with sparsity=.05) to test L3D's ability to resolve models with superimposed features.

\subsubsection{Decomposition}

\begin{figure*}
  \centerline{\includegraphics[width=\textwidth]{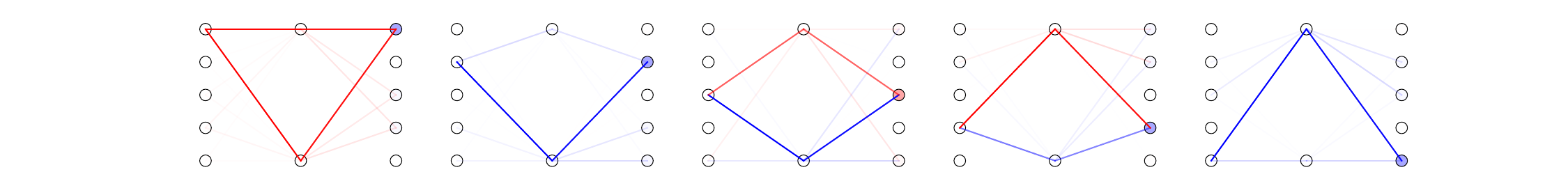}}
  \centering
  \caption{L3D subnetwork decomposition of TMS. Each subnetwork corresponds to the encoder/decoding of a single input feature.}\label{fig:3_tms_subnetworks_first5}
\end{figure*}

\begin{figure}
  \centerline{\includegraphics[width=\columnwidth]{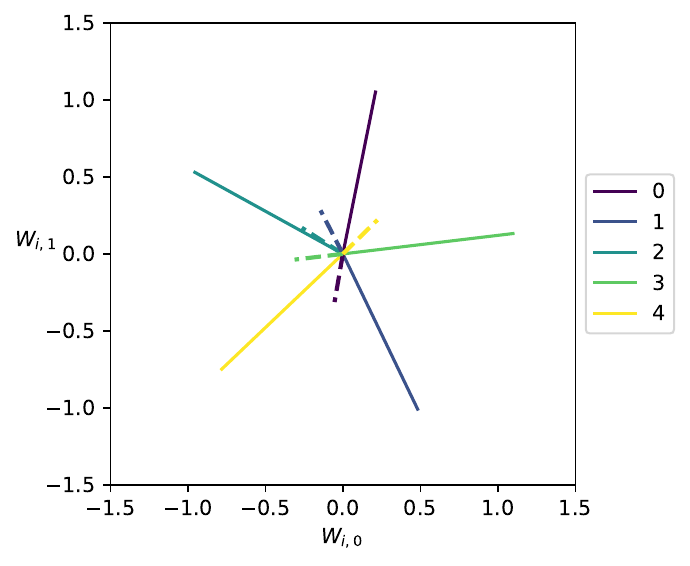}}
  \centering
  \caption{The encoder/decoder directions of the original model (solid lines) and each subnetwork (dashed lines). The directions learned by each subnetwork are nearly perfectly parallel to the encoding for each input feature. The colors of the lines refer to the input index each embedding represents.}\label{fig:s2_tms_encoder_directions}
\end{figure}

We decomposed the TMS model into 5 subnetworks, using rank-1 parameter tensors. L3D successfully decomposed the model into subnetworks corresponding to the encoding and decoding of each feature (a $X_i:\hat{X}_i$ circuit). Figure \ref{fig:3_tms_subnetworks_first5} shows the decomposition. Moreover, the encoder directions of the learned subnetworks are nearly perfectly parallel to the original embedding of each input index (Figure \ref{fig:s2_tms_encoder_directions}). One thing to note is that parameter vectors do not have a preferred direction. L3D is equally likely to identify a parameter vector in the direction of $\theta$ as it is in the direction of $-\theta$. This is why, for example, the weights in subnetwork 1 are in the opposite direction as the original network (Table \ref{tab:toy_models}).

This decomposition resulted in a reconstruction error of 19\%. The reconstruction error is related to the interference between features when multiple features are active in the same sample. We expect decompositions of higher dimensional networks to exhibit less reconstruction error, as the amount of nearly orthogonal parameter vectors (non-interfering) that can be compressed into parameter space scales exponentially with dimension. We see this effect in the next higher-dimensional toy model where the reconstruction loss is in fact lower. 

\subsubsection{Intervention}

\begin{figure}
  \centerline{\includegraphics[width=\columnwidth]{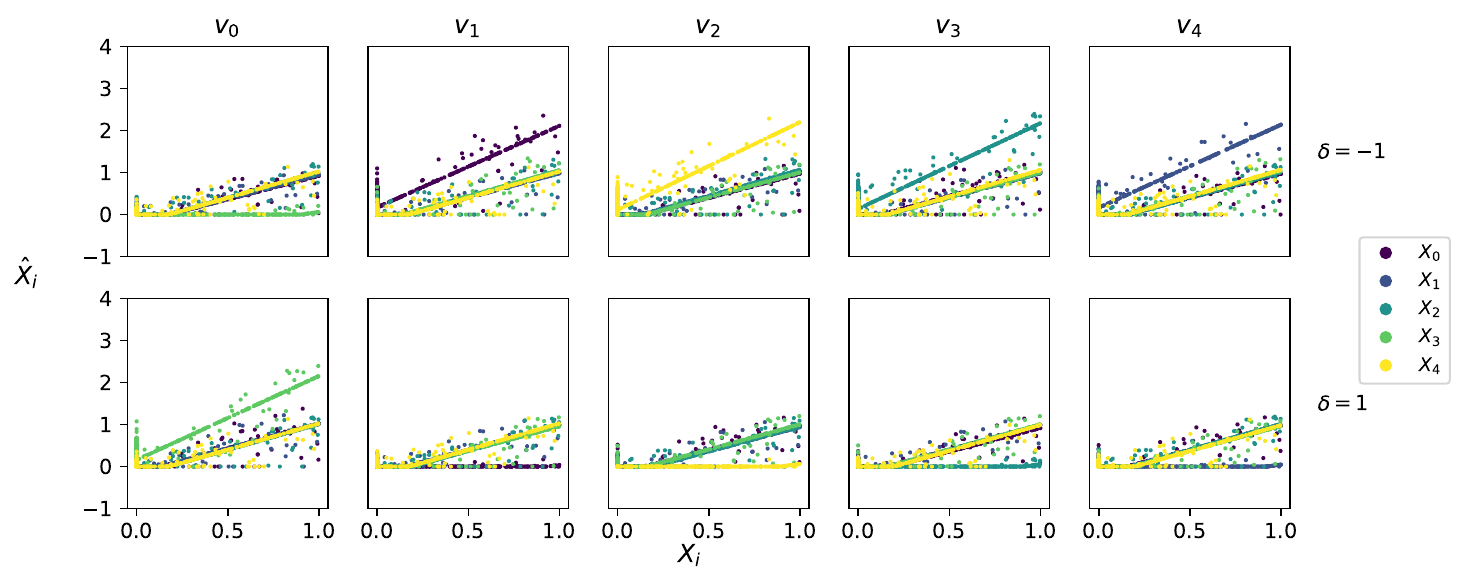}}
  \centering
  \caption{The effect of intervening on the TMS model in the direction of each subnetwork. We generated 1000 inputs from the TMS input distribution (x-axis), intervened on each subnetwork $v_i$ with magnitude $\delta$ and measured the change in outputs (y-axis) for each sample. The outputs corresponding to the index relevant to each subnetwork experienced the most change.}\label{fig:4_tms_intervention}
\end{figure}

\begin{figure}
  \centerline{\includegraphics[width=\columnwidth]{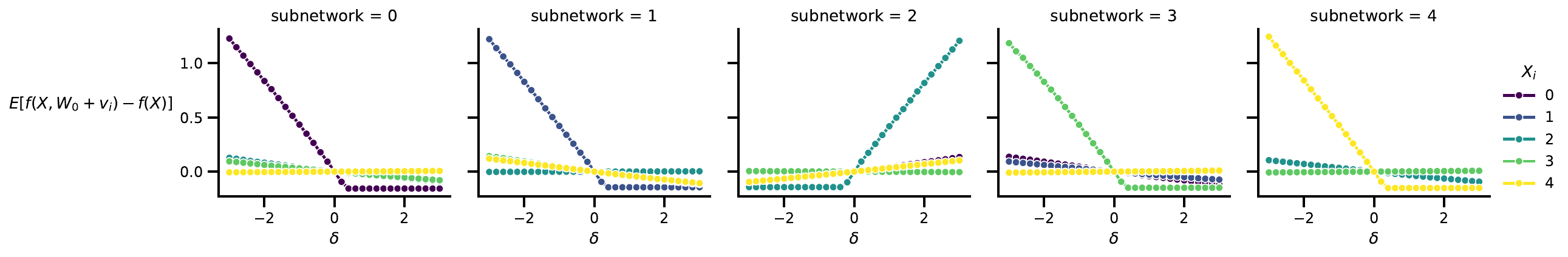}}
  \centering
  \caption{The effect of intervening at various values of $\delta$ in the direction of each subnetwork. The y-axis represents the average amount an output changed (data points colored by output index), when perturbed an amount $\delta$ in the direction of a subnetwork. }\label{fig:4_tms_intervention_mean}
\end{figure}

Parameter vectors learned by L3D can be used to intervene on model behavior. In principle, we could finetune a model using only selected subnetworks (See \ref{subsec:extensions}). While we did not go the extent of finetuning a model, we explored the effect of perturbing a model's parameter space in the direction of a subnetwork (by an increment of $\delta$), as described in Section \ref{subsec:intervention}. If subnetworks do in fact represent sparse computations, we hope that intervening on a subnetwork has a strong effect on the predictions of relevant samples, and little effect on others. As shown in Figures \ref{fig:4_tms_intervention} and \ref{fig:4_tms_intervention_mean}, moving the TMS model in the direction of a single subnetwork did in fact achieve this. Perturbing in the direction of subnetwork 1 primarily affected samples where feature 1 was active, with a small effect on the inputs that had interference with feature 1's embeddings. In fact, for TMS, we could successfully fully ``turn off" a computation by moving far enough in the direction of the subnetwork. (Although for models with more complex loss landscapes, ``turning off" a computation is not as straightforward, as we will later discuss).

\subsection{Toy Model of Circuit Superposition}

\subsubsection{Setup}

TMS exhibits \textit{feature superposition} - the input features' low dimensional embeddings are non-orthogonal. However, the sparse \textit{circuits} in the original TMS we decomposed are notably \textit{not} in superposition - a given weight or parameter is only relevant for a single circuit and circuits. It seems highly unlikely that real world model circuits would decompose this way, since learning circuits composed of perfectly orthogonal parameter vectors limits the amount of circuits that can be contained in a given set of parameters. We therefore developed a toy model of \textit{circuit superposition} (TMCS) in order to analyze L3D's ability to resolve such circuits. We define \textbf{circuit superposition as a phenomenon by which subnetworks share parameter elements, and even more generally have non-orthogonal parameter vectors.}

Our toy model of circuit superposition (Toy model 2 in Table \ref{tab:toy_models}) uses the same architecture and input data distribution as TMS, but is trained to predict \textit{linear combinations of the input features }($X \mapsto A X$). We set the entries of $A$ as uniform random values between 0 and 3 (chosen arbitrarily) and generate input-output pairs to train the toy model with. We used an model with 10 inputs, 5 hidden layers, and 10 output features (although such a model does not need to have the same number of input and outputs).

If subnetworks are only relevant to a small set of inputs, then we would expect each subnetwork to compute an input feature's contribution to the output vector. If this is the case, then individual parameters would be involved in multiple subnetworks: ${W^{dec}}_{i,1}$ (the set of parameters connecting the hidden nodes to the first output node) will contain information about both $A_{1,1}, A_{2,1}, A_{3,1}...$. Put another way, the subnetworks will interfere with each other - parameter directions associated with each subnetwork will be non-orthogonal. 

\subsubsection{Decomposition}

\begin{figure*}[htbp]
  \centerline{\includegraphics[width=\textwidth]{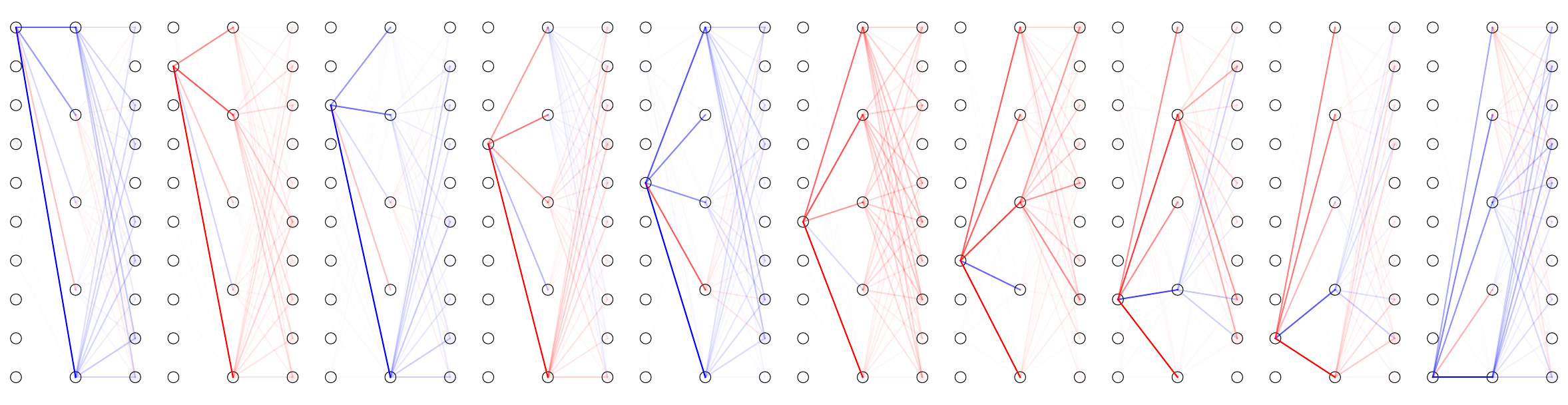}}
  \centering
  \caption{The subnetworks L3D successfully decomposes the TMCS model into subnetworks computing the contributions of each input feature to the output vector.}\label{fig:5_circuit_superposition_decomposition}
\end{figure*}

We decomposed TMCS into 10 subnetworks of rank-1 parameter tensors (Figure \ref{fig:5_circuit_superposition_decomposition}) with a reconstruction loss of 6.4\%. The subnetworks each strongly corresponded to a single input feature, as we expected.

\begin{figure}[htbp]
  \centerline{\includegraphics[width=\columnwidth]{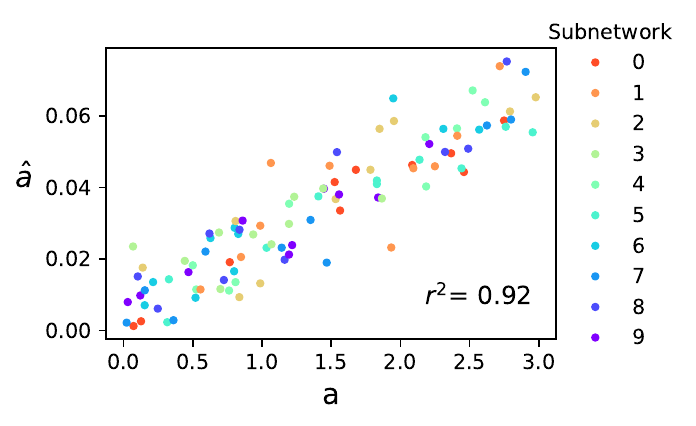}}
  \centering
  \caption{We use the L3D subnetworks to derive the values of $A$ and compare them to the to true coefficients used to train TMCS. We see that they have a very high correlation.}\label{fig:6_circuit_superposition_coefficients}
\end{figure}

Since each subnetwork theoretically corresponds to the contributions of a single input feature, we should be able to reconstruct the original $A$ values from each subnetwork. To derive $A$ from each subnetwork, we (1) identified the which column in the subnetwork's $W^{dec}$ direction has the largest norm and then (2) traced the weights of the network through that path. That is for subnetwork $k$: 
\begin{align}
  &j^* = \underset{j}{\text{argmax}} ||{{{W^{dec}}_j}_k}||_2 \\
  &\hat{a}_{i,j^*} = {{{W^{enc}}_{i,j^*}}}_k {{W^{dec}}_{i,j^*}}_k \notag
\end{align}
Recall the parameter vectors are normalized to be unit vectors so we expect them to be a scalar multiple of the true $A$ values. As seen in Figure \ref{fig:6_circuit_superposition_coefficients}, our derived $\hat{a}$ had a very high correlation to the original $a$ values ($r^2 = 0.92$).

\subsection{Higher Rank Circuits}

\subsubsection{Setup}
Because each subnetwork in TMCS traces the path of a single input neuron, the underlying subnetworks should inherently have a rank of 1. In order to test the ability of L3D to learn higher rank circuits, we developed a toy model with inherently higher rank circuits. For this model, we used the same set up as TMCS, but we correlated the sparsities of sets of input features. We used 30 input features, and we filtered our data to ensure that input features 1-5, 6-10, etc, are always active ($>0$) or inactive ($<0$) together. In this setup, circuits should always be associated with groups of 5 input features and so should have a rank of 5 (diagram shown in Figure \ref{fig:s1_high_rank_circuit_setup}). 

\subsubsection{Decomposition}

\begin{figure*}[htbp]
  \centerline{\includegraphics[width=\textwidth]{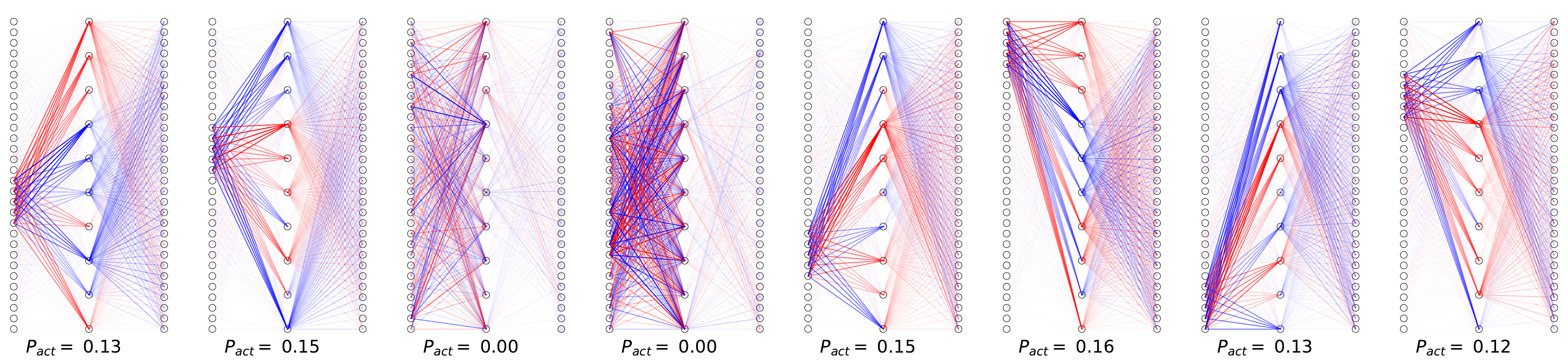}}
  \centering
  \caption{Parameter representations learned by L3D for the high rank circuit decomposition task. Each subnetwork corresponds to a correlated group of feature. The third and fourth subnetworks are ``dead" subnetworks that did not make it into the $\text{topK}$ selection at all during the final epoch.}\label{fig:7_high_rank_decomposition}
\end{figure*}

Although we expect the model to have 6 subnetworks, we used excess parameter tensors ($n_v=8$) in order to allow more flexibility in learning. We tracked the fraction of inputs for which a subnetwork was used in the $\text{topK}$ reconstruction ($P_{act}$) to identify which were ``dead subnetworks", and report $P_{act}$ from the last epoch. Furthermore, although we expected the underlying subnetworks to be rank 5, we experimented with using different rank representations to see how well lower-rank parameter directions could represent the model. Interestingly, rank-1 representations of the parameter tensors were able to represent the model nearly as well as rank-5 representations (Figure \ref{fig:s5_high_rank_circuits_loss_vs_rank}). In Figure \ref{fig:7_high_rank_decomposition}, we show the decomposition of a rank-3 decomposition. L3D successfully learned a subnetwork corresponding to each of the 5 input feature groups, as well as a number of dead circuits. The higher and lower rank decompositions also learned similar subnetworks (Figure \ref{fig:s6_high_rank_decompositions}). When we trained L3D without these additional subnetworks, the reconstruction loss often got caught in local minima. Similar to training sparse autoencoders \cite{cunningham2023sparse}, having extra degrees of freedom allows for better learning, even if at the end of training the extra subnetworks are never active.

\subsection{Toy model with Complex Loss Landscape}

\subsubsection{Setup}

In the previous models, other than the ReLU discontinuity the model's were linear transformations between inputs and outputs. We should expect their loss landscapes to therefore be well-behaved, with local attributions being perfectly representative of global attributions (up until the ReLU discontinuities). However, we wanted to test the limitations of a L3D on a model with a more complex loss landscape, especially when it comes to intervening with a subnetwork.

We therefore trained a multi-layer model to predict multiple non-linear functions of input features at once. We trained a GeLU network for $X_i \mapsto X_i^2$. We used a network with 4 hidden layers of 10 neurons each, and 5 input and output neurons. Once again, the input features are sparse (and range from -1 to 1), incentivizing the toy model to learn circuits in superposition whose interferences will cause minimal errors on the sparse input distribution. 

We a priori expected the model to have 5 subnetworks, one for each input feature. Although it is less clear what rank the tensors of the underlying circuits should be, there are not inherent reasons to believe subnetworks should be low rank the way there was in the TMS model. 

\subsubsection{Decomposition}

\begin{figure*}[htbp]
  \centerline{\includegraphics[width=\textwidth]{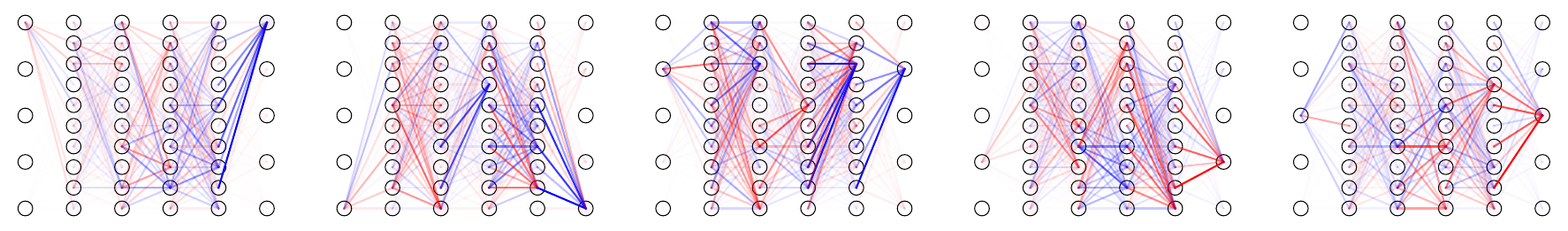}}
  \centering
  \caption{Subnetworks learned by L3D for the $X \mapsto X^2$ model. Each subnetwork is relevant to a single input/output index.}\label{fig:8_squared_subnetworks}
\end{figure*}

To allow for slightly higher rank subnetworks but still compress the dimensions of the model, we decomposed our model into 5 rank-2 parameter tensors. Additionally, instead of varying rank, we experimented with using different numbers of subnetworks to represent our model. In the 5-subnetwork decomposition (Figure \ref{fig:8_squared_subnetworks}), we found subnetworks tracing the path of $X_i \mapsto X_i^2$ for each index $i$. However, this decomposition had a relatively high reconstruction error of 32\%. Much of this was probably because we kept our $\text{topK}$ hyperparameter constant (at $k=0.1$) throughout all our our models for consistency. With only 5 subnetworks, this means that each sample's reconstruction will use $<1$ subnetwork on average, limiting the minimum reconstruction error the network can achieve. 

We also experimented with holding rank constant (we dropped to rank 1 for this) and decomposed the model into different numbers of subnetworks (3, 5, 10, and 15 subnetworks). In our 3-subnetwork decomposition, L3D still learned subnetworks corresponding to single input features, but can of could only represent 3 out of the 5 inputs. As we added more subnetworks, L3D was able to successfully learn more expressive decompositions of the model that reduced reconstruction error (Figure \ref{fig:s11_squared_features_vs_loss}). Each decomposition continued to learn subnetworks specific to a single input/output index, with the larger decompositions resulting in a few more dead subnetworks as well (Figure \ref{fig:s10_squared_decompositions_features}).

\subsubsection{Intervention}

\begin{figure*}[htbp]
  \centerline{\includegraphics[width=\textwidth]{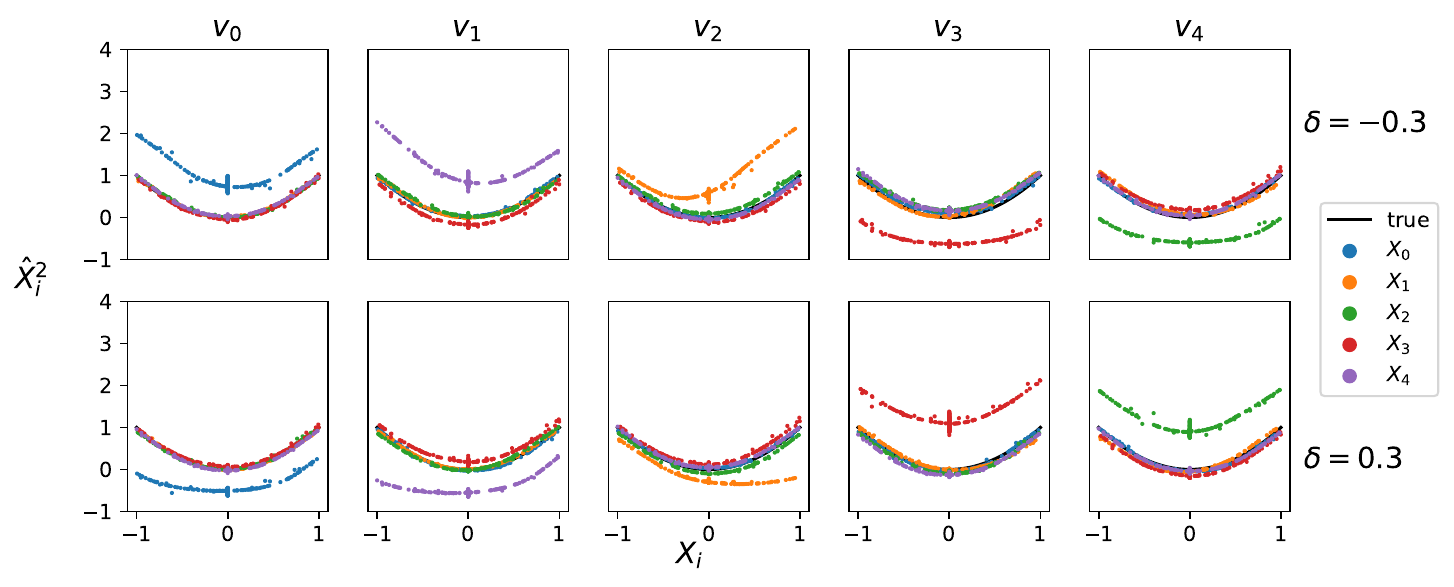}}
  \centering
  \caption{The effect of intervening on each subnetwork in the $X \mapsto X^2$ model. We generated 1000 inputs from the TMS input distribution, intervened on each subnetwork with magnitude $\delta$ and measured the change in outputs for each sample. Only the outputs that involve each subnetwork effectively changed.}\label{fig:9_squared_intervention}
\end{figure*}

Intervening on these circuits helps us understand how much local loss landscape is representative of global loss landscape, particularly when it comes to inactive subnetworks remaining inactive as we move through parameter space. If local loss landscape is truly representative of global loss landscape in this way, then intervening on on a single subnetwork should result in only the set of samples that relies on the subnetwork, even if we move very far in that direction. Figure \ref{fig:9_squared_intervention} shows our results for these interventions on the $X \mapsto X^2$ model. Even in this more complex toy model, local loss landscape is a relatively good approximation of the global loss landscape. We can move our model parameters in a direction of interest and have a large impact on the predictions of the relevant inputs and a minor impact on others. If we perturbed far enough (Figure \ref{fig:s7_squared_intervention_more_deltas}), we did begin to see effects on the predictions of other samples, but the ratio of change in predictions to the relevant samples to those of the irrelevant samples remains very high.

Figure \ref{fig:9_squared_intervention} shows changes in predictions as we move in a single direction in parameter space. We also wanted to understand how subnetworks might interact with each other as we move through parameter space. In Figure \ref{fig:s9_squared_intervention_multi_features} we perturbed multiple subnetworks at once, and measured the new predictions. For the most part, the subnetworks had little inference with each other: the relevant output values for each subnetwork moved relatively independently of each other. 

\subsection{Real world models}

Finally, to show the promise of this method to pull out relevant features from real world models, we used L3D to decompose blocks of a language model and computer vision model. These results are primarily qualitative, and were run with minimal compute and little hyperparameter tuning. Our choices for model block, number of subnetworks, and subnetwork rank were relatively arbitrary. 

These models do not have well-characterized subnetworks the same way our toy models do. To briefly analyze the function of the subnetworks we identified, we looked at the top samples that each subnetwork is relevant to. We computed this metric as described in Section \ref{subsec:intervention}. We also computed the most affected logits, for each of the top samples ($x_i$) for each subnetwork ($v_i$):
\begin{equation}
\text{argmax}[\text{abs}(\nabla_\delta f(x_i, W_0 + \delta v_j) |_{\delta=0})]
\end{equation}
Because these models output probabilities, we used KL-divergence as our divergence metric when performing L3D for these models.

\subsubsection{Language Model}

We decomposed attention block 7 of the tiny-stories-8M model into 100 subnetworks. We used ranks that are approximately 1/10 the original dimensions of the network (see Section \ref{sec:realworld_hyperparams}), and once again use $k=.1$. Although L3D can decompose any number of parameter blocks, or all parameters in a model into subnetworks, we limited L3D to a single block to keep memory and compute time low. We chose an attention block because this has been a challenging component of a transformer for SDL to extract features from \cite{sharkey2025open}. We chose a middle layer of the model so that we identify subnetworks that are neither so high-level that they perfectly line up with next-token prediction, and not so low-level that they perfectly line up with token id.

Table \ref{tab:transformer_fav} shows the top samples of 10 cherry-picked circuits, and Table \ref{tab:transformer_full} shows the top samples for all of the circuits. Although L3D had a relatively high reconstruction error at 40\% (potentially due to only using 100 subnetworks), subnetworks seemed relatively interpretable. Even in the full set of circuits, most corresponded to a human-interpretable computation, such as detecting word pairs and phrases, certain parts of grammar, subjects from previous parts of a sentence. We leave it as an exercise to the reader to annotate and interpret each circuit. 

\subsubsection{Computer Vision Model}
\begin{figure*}[ht]
  \centerline{\includegraphics[width=.8\textwidth]{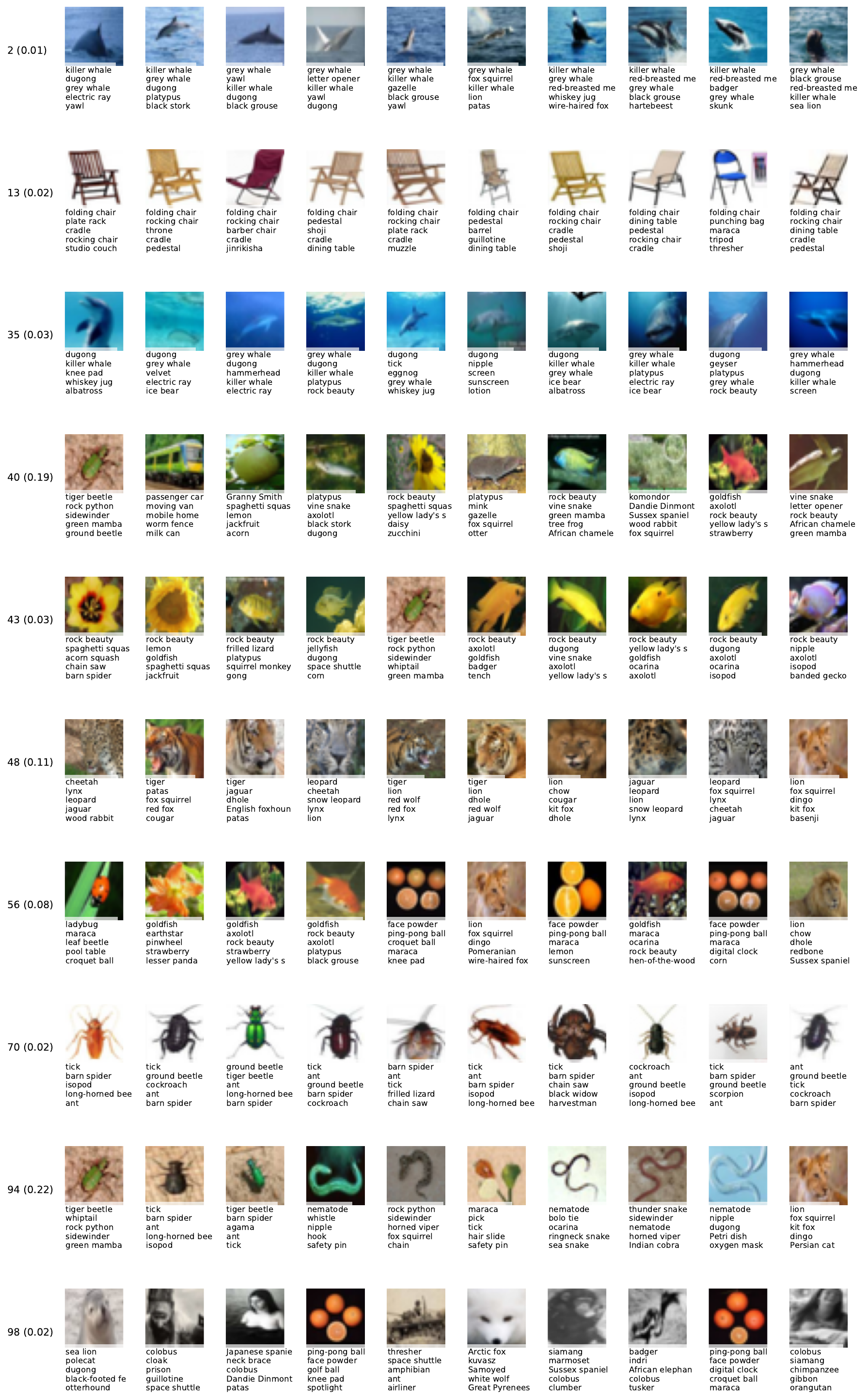}}
  \centering
  \caption{For 10 of our favorite subnetworks in the mobilenet-v3-small decomposition, we computed the top most affected samples (images). For each of those samples, we computed which logits had the highest gradient with respect to the subnetwork direction.}\label{fig:favorite_cnn_circuits}
\end{figure*}

We decomposed convolutional block 4 of the mobilenet-v3-small model. Once again we choose a middle layer such that the subnetworks we identify are neither involved in low-level computations that would likely require additional pixel attribution methods to interpret, or so high level they perfectly line up with classification. We used similar hyperparameters as in the transformer decomposition, as described in \ref{sec:realworld_hyperparams}. Once again, we computed the top most affected samples for each circuit. We show the samples for 10 cherry-picked circuits in Figure \ref{fig:favorite_cnn_circuits} and for all 100 circuits in Figure \ref{fig:cnn_full_circuits}. Some types of computations include recognition of certain animal faces, colors, backgrounds, and objects. Interestingly, although L3D's decomposition of mobilenet-v3-small had a lower reconstruction error (23\%), many of the subnetworks initially seem somewhat less human-interpretable. We suspect doing pixel attribution may help resolve some of the subnetwork computations as subnetworks might be picking out specific shapes and forms that are not obvious from just viewing the subnetworks most relevant samples at a high level. 

\onecolumn
\footnotesize	
\begin{longtable}{|p{0.15\textwidth}|p{0.45\textwidth}|p{0.35\textwidth}|}
\hline
\textbf{Id ($P_{act}$)} & \textbf{Input Text} & \textbf{Top Logits} \\
\hline
\endfirsthead 
\hline
\textbf{Id ($P_{act}$)} & \textbf{Input Text} & \textbf{Top Logits} \\
\hline
\endhead
\hline
\multicolumn{3}{r}{\textit{Continued on next page}} \\
\hline
\endfoot

\hline
\endlastfoot
& & \\
\multirow{5}{*}{\textbf{0 (0.072)}} & be more careful when eating spicy food. From that & day, day, Monday, side, night \\
& too because she helped the bird. From that & day, side, umm, ts, Balls \\
& she should have been more careful. From that & day, day, side, cers, ts \\
& tummy hurt. From that & day, side, Balls, acas, ters \\
& . From that & day, side, acas, cers, Balls \\
& & \\
\multirow{5}{*}{\textbf{5 (0.107)}} & together.Once upon & a, an, SEC, irled, clip \\
& best friends.Once upon & a, an, SEC, clip, irled \\
& , so they stay colorful and clean."Once upon & a, an, orse, ship, ream \\
& you for being so persistent, daddy."Once upon & a, an, ud, orse, SEC \\
& became good friends.Once upon & a, an, clip, SEC, irled \\
& & \\
\multirow{5}{*}{\textbf{16 (0.028)}} & it first!" Sara says. "We want to see the treasure!" & Ben, Tom, She, she, Tim \\
& . They are not ours to take. They are the sea's to give." & They, Tom, Ben, Mom, Tim \\
& race!" Ben said. "I bet I can go faster than you!" & He, Lily, Mia, , he \\
& is not good to touch. Mom said some mushrooms are bad." & But, Mom, They, Ben, Lily \\
& chicken too. They are all good for you." & They, Mom, , The, Lily \\
& & \\
\multirow{5}{*}{\textbf{18 (0.060)}} & . It was your treasure." Ben shook his & head, izing, Warning, iated, alking \\
& . Lily and Ben look at each & other, enlarged, OUT, pping, heit \\
& at the shell. They looked at their mom. They looked at each & other, wait, pace, lower, bribe \\
& clumsy, Sam," Tom said, shaking his & head, neck, chin, heads, eyebrows \\
& chicken too. They are all good for you." Tom shook his & head, Warning, FUN, izing, Save \\
& & \\
\multirow{5}{*}{\textbf{21 (0.094)}} & dad were hurt too. They went to the & hospital, doctor, nurse, car, pool \\
& They hide the letter under the & couch, bed, sofa, table, slide \\
& They could play on the & swings, beach, subway, climbers, Safari \\
& the old lady talked on the & phone, telephone, cellphone, plaza, cafeteria \\
& to see who could get the best score. Tim threw the & ball, balls, basketball, trash, seeds \\
& & \\
\multirow{5}{*}{\textbf{30 (0.048)}} & She did not see her & ., and, feet, hand, Mom \\
& in the bathtub. She did not hear her & Mom, voice, mother, big, brother \\
& She said to her & ,, daughter, little, friend, Mom \\
& outside.  Lily told her & mom, ,, grandma, Mom, that \\
& night. One day, she told her & friend, friends, Mom, parents, mother \\
& & \\
\multirow{5}{*}{\textbf{59 (0.023)}} & to sleep." Tom gave back the jewelry and said, "Thank & you, background, ptions, mats, react \\
& Lily nodded and said, "Thank & you, opes, ptions, mats, speakers \\
& , "Thank & you, ptions, background, technique, bolts \\
& It looked happy. "Thank & you, ptions, opes, bolts, zel \\
& Ben smiled and said, "Thank & you, ptions, opes, background, bolts \\
& & \\
\multirow{5}{*}{\textbf{71 (0.080)}} & angry. Lily and & Ben, Tom, Jill, Mint, Fay \\
& ," Tom said. Lily and & Tom, itt, est, hy, ippers \\
& It had a cut on its leg. Lily and & Ben, Tom, Mint, Flor, Shawn \\
& Anna and & Ben, iner, ability, astical, sub \\
& Lily and & Ben, Tom, Jack, Mark, Peter \\
& & \\
\multirow{5}{*}{\textbf{76 (0.411)}} & They like to play with their toys and books & in, and, ,, together, ," \\
& day, Timmy went to play with his friends in the park & ,, and, with, again, for \\
& . Max loved to play with his friends at the park & ,, every, and, because, with \\
& are friends. They like to play in the park & with, and, every, near, , \\
& had a big toy that she really wanted & to, ,, and, !, but \\
& & \\
\multirow{5}{*}{\textbf{86 (0.110)}} & proud of herself for helping her furry friend.Once upon a time & there, at, in, later, it \\
& listen to her mom and always be safe.Once upon a time & there, in, at, it, they \\
& under her plate or give them to the dog. One day & she, the, when, they, her \\
& friends. They played together every day. One day & the, it, they, Tim, Tom \\
& importance of sharing and being kind to his friends.Once upon a time & there, at, in, later, with \\
\caption{For 10 of our favorite subnetworks, we computed the top most affected tokens, in terms of their KL-divergence compared to several reference outputs on the next-token prediction task. For each of the texts, the last token is the token that was found to be the most affected for each subnetwork. For each top token, we also computed the logits with the highest absolute gradients with respect to the subnetworks..}\label{tab:transformer_fav}
\end{longtable}
\twocolumn
\normalsize

The decomposition for mobilenet-v3-small also had much higher numbers of dead circuits (40\%). We suspect adding an auxiliary loss term as in \cite{gao2024scaling} might help alleviate this issue as well as improve reconstruction loss further.

\section{Discussion}

L3D is one of the earliest parameter-based decomposition methods. For this reason, we have focused our work on demonstrating the fundamentals of L3D on toy models, and showcasing its promise with more complex models. Here we discuss what we believe are simple improvements to L3D that could enhance its performance and real-world use cases to which to extend L3D. Finally, we discuss unresolved challenges and limitations of L3D. 

\subsection{Simple Improvements}
In this work, we did not focus on optimizing L3D, and we chose nearly identical hyperparameters for all of our decompositions. 

\textbf{Hyperparameter Choice}: For all of our toy model decompositions, we always chose our $\text{topK}$ hyperparameter as $k=0.1$, even when it was clear that certain toy models should have larger numbers of subnetworks activated per sample than others (For example, the $X \rightarrow X^2$ model with 5 inputs and 5 outputs, decomposed into 5 networks, should probably have $k \geq 0.2$). Too low of $k$ choice is likely responsible for the high reconstruction loss of some of our models. Similarly, we chose the ranks of the subnetworks somewhat arbitrarily. Some preliminary research aims to understand the relationship between rank, compressibility, and interference of subnetworks \cite{hanni2024mathematical,bushnaq2024circuits}, and a better understanding of this relationship could help us choose better hyperparameters for L3D. 

\textbf{Scaling up}: Naturally, the most exciting applications of L3D are with real-world models. While we briefly shared some results on larger models in order to demonstrate L3D's promise, we by no means did a deep dive into the results. We think L3D can be scaled up to real-world models and can help answer open questions related to the amount of superposition present in different blocks of models, how circuits and features interact with each other and which parts of a model's architecture are the most over- or under- parameterized. 

\subsection{Extensions}\label{subsec:extensions}

There are a also some higher effort extensions to L3D that may give it more real-world relevance. 

\textbf{Finetuning}: Our intervention experiments showed promise that subnetworks of L3D could be perturbed in ways that only affect the predictions of relevant samples. As we describe in Section \ref{subsec:extensions}, this could be taken one step further by finetuning a model on a specific set of parameter directions. Using L3D networks, we could finetune a model on a specific set of parameter directions identified by L3D by freezing the current set of weights and learning an adapter consisting of linear combinations of the subnetworks of choice. This could also benchmark the intervention capabilities of L3D versus other mechanistic intervention strategies such as SDL-derived steering vectors. For example, we might use L3D to identify various subnetworks involved in sycophancy, refusal, and other undesired behaviors. After collecting curated data with the goal of finetuning away such behaviors, we could finetune L3D only in the direction of the behavior-related subnetworks and test how well the model achieves our desired output compared to other intervention strategies. 

\textbf{Identifying Specific Circuits with Contrastive Pairs}:
We developed this method as an unsupervised decomposition method, with goals comparable to those of SDL. However, the methods of L3D could be easily modified to use supervised signals to identify specific circuits of interest. Rather than using gradients of divergence of random pairs, we could decompose gradients of divergence between curated pairs of samples that isolate a behavior of interest, or between outputs of different models on the same sample.

\subsection{Challenges}
Although many of the improvements and extensions of L3D are highly addressable, we think there are some fundamental challenges with parameter-based decomposition methods that may not be easily resolved.

\textbf{Local Attribution}: L3D's algorithm hinges on the somewhat surprising phenomenon that local gradient approximations work reasonably well as attribution methods. They clearly work well in the toy models we used for L3D and at least demonstrate promise for the circuits we found in our real world models. However, do they work for all circuits? In our work, we use a randomly selected sample to be our "reference" output with which to compute divergence gradient. By using a randomly selected sample, rather than a single ``reference output" such as the mean of the output distribution, we hope that the random noise in the reference sample will average out the effects of any non-convexity in the loss landscape. However, perhaps even in this setup there are parameter directions that are highly non-convex on which it will be difficult to perform local attribution. Quantifying different types of ``dark matter" of parameter decomposition by analyzing reconstruction loss \cite{engels2024decomposing} could better help us characterize these limitations.

\textbf{Relationship to overparameterized models}: Going one step further, we suspect that the reason local attribution methods work so well is because large models are probably overparameterized \cite{kawaguchi2016deep,choromanska2015loss,dauphin2014identifying,soudry2017exponentially}. Larger models may have wider loss basins, or more degeneracies near their global minima \cite{keskar2016large,sagun2017empirical}, making local attribution methods less likely to break down as we move through parameter space. If in the future, a learning algorithm is developed that has fundamentally different limitations that stochastic gradient descent and its relatives, we might lose this property. Moreover, circuit activations might no longer be sparse. A new learning process might be able to compress subnetworks in such a way that subnetworks have very high levels of interference with each other - removing the degeneracy assumption that underlies L3D. 

\textbf{Interpretation of a circuit}: Finally, we should address the definition of ``circuits". It is still not well agreed upon what a ``feature" is in relation to large networks, and the definition of what should constitute a circuit or subnetwork is even less clear. Is our definition of a circuit - sparsely active subnetworks that can move outputs within the original output distribution - too restrictive? If there is a circuit that is relevant to every output, a sort of ``scaffolding" for more specific circuits - should it be included in the decomposition? If, after identifying the structure of subnetworks, we cannot interpret it beyond a description of its end results, are circuits any more informative than the features they are computing? If parameter decomposition is a viable strategy for understanding and intervening with large networks, these questions will be important for the mechanistic interpretability community to address.

\onecolumn
\section{Acknowledgments}

Thank you to Daniel Filan and Dan Braun for additional comments and feedback. This work was funded by Open Philanthropy and the Machine Learning Alignment and Theory Scholars program (MATS). 

\section{Code Availability}

Code for this project can be found at  https://github.com/briannachrisman/eigenestimation.

\clearpage


\clearpage

\bibliography{writeup.bib}
\bibliographystyle{icml2025}

\newpage
\appendix
\renewcommand{\thefigure}{S\arabic{figure}} 
\renewcommand{\theHfigure}{S\arabic{figure}} 
\setcounter{figure}{0} 
\onecolumn

\section{Definitions}

\subsubsection{Dimensions}
$n_s$: The number of samples in a batch of inputs 

$n_i$: The dimensions of a single input vector to a model

$n_o$: The dimensions of a single output vector from a model

$n_w$: The number of parameters values in a model.

$n_v$: The number of subnetworks or parameter directions chosen to decompose a model.

\subsubsection{Model Syntax}
$X \in \mathbb{R}^{n_s \times n_i}, x \in \mathbb{R}^{n_i}$: Batch and individual input vectors to a model.

$y_r \in \mathbb{R}^{n_o}$: A reference output vector

$W \in \mathbb{R}^{n_w}, w \in \mathbb{R}$: The set of and individual parameter values of a model

$f: \mathbb{R}^{n_s \times n_f} \mapsto \mathbb{R}^{n_s \times n_o}$: A model mapping a set of input vectors to a set of output vectors.

$f(X, W)$: The output of model $f$ with parameter values $W$ on input $X$.

$f(X, W_0)$ or $f(X)$: The output of model $f$ with fixed parameter values $W_0 $. $W_0$ is the set of learned parameter values from model training.

$D$: Divergence metric between two vectors. Typical divergence metrics are mean-squared error for regression-type outputs, and KL-divergence for probability-type outputs. 

\subsubsection{Decomposition Syntax}
$V$ (or $V^{out}) \in \mathbb{R}^{n_v \times n_w}, v$ (or $v^{out}) \in \mathbb{R}^{n_w}$: The set of or individual parameter directions that are used to decompose a model. $V^{out}$ can be used to transform parameter directions in the subnetwork vector space back into the original parameter space of the model. 

$V^{in} \in \mathbb{R}^{n_w \times n_v}$: Transforms the original parameter space of the model into the subnetwork vector space. 

$r$: The rank of each component of the decomposition vectors corresponding to tensors in the original model. 

\subsubsection{Training}

$L$: The L2 reconstruction loss used to optimize $V^{in}$ and $V^{out}$.

\subsubsection{Measuring and Intervention}
$I(x_i, y_j, v_k)$: The impact of subnetwork $v_k$ on the divergence between sample outputs $f(x_i)$ and $y_j$, or averaged across many $y_j$ reference outputs.

$\delta$: A scalar value to move $W$ in a specific direction.

\section{Additional Methods}

\subsection{Low-Rank Tensor Representation}\label{sec:low_rank}

We use low-rank representations of our $V^{in}$ and $V^{out}$, and correspondingly learn low-rank circuits.

While $W$ is a vector containing all model parameters, these parameters are typically organized into tensors, $W=\{w_i\}_i$.

If our parameters are structured as tensors $W = \{w_i\}_i$, each subnetwork or parameter component can be expressed as $V^{in}_i = \{\{v^{in}_i\}\}_i$ and $V^{out}_i = \{\{v^{out}_i\}\}_i$, where each component corresponds to a specific tensor in the original model parameters. To ensure that each of these tensors remains low-rank, we employ the \textbf{Tucker decomposition} \cite{tucker1966some} (a method for factorizing high-dimensional tensors into a core tensor and a set of factor matrices):

The Tucker decomposition decomposes a tensor \( \mathcal{v} \in \mathbb{R}^{I_1 \times I_2 \times \dots \times I_N} \) into a core tensor \( \mathcal{G} \) and a set of factor matrices \( \mathbf{U}^{(n)} \):
\begin{equation}
  \mathcal{v} \approx \mathcal{G} \times_1 \mathbf{U}^{(1)} \times_2 \mathbf{U}^{(2)} \dots \times_N \mathbf{U}^{(N)}
\end{equation}

where:
- \( \mathcal{G} \in \mathbb{R}^{R_1 \times R_2 \times \dots \times R_N} \) is the core tensor capturing interactions between modes.
- \( \mathbf{U}^{(n)} \in \mathbb{R}^{I_n \times R_n} \) are the factor matrices, representing a low-rank basis along each mode.
- \( \times_n \) denotes the n-mode product of a tensor with a matrix.

\subsection{Toy Model Training}\label{sec:toymodel_hyperparams}

For all of our toy models (except the $X \mapsto X^2$ model), we generate uniformly random inputs between 0 and 1. For $X \mapsto X^2$, we generate uniformly random inputs between -1 and 1. For all toy model data, we use a sparsity value of sparsity=.05. We generate 10000 datapoints and train for 1000 epochs with a batch size of 32. We use an AdamW optimizer with a learning rate of 0.001. 

\subsection{L3D Toy Model Training}\label{sec:L3D_hyperparams}

To train L3D for the toy models, we use the same training distributions as in each toy models. Although optimal hyperparameter values probably depend on the model size, and the rank and number of parameter tensors, we use the same hyperparameters for all of our models. We generate only 1000 datapoints, with a batch size of 32, and train for 1000 epochs. We use an AdamW optimizer with a learning rate of 0.01, and a learning decay rate of .8 every 100 steps. We always use a $\text{topK}$ hyperparameter of $k=0.1$. We include all of the model's parameter tensors, including biases, in the decomposition. 

\subsection{L3D Real World Model Training}\label{sec:realworld_hyperparams}

To decompose tiny-stories-8M, we train L3D using 10000 16-token texts randomly sampled from the tiny-stories dataset. For mobilenet-v3-small, we train L3D using 10000 images samples from CIFAR-100.

For both our models, we train for 100 epochs with a learning rate of .005 and a decay rate of .8 every 10 epochs. We computed the top samples using 10000 additional randomly generated images/texts from the same distribution as training, and averaging the contribution of each subnetwork to each sample across 10 reference outputs. 

For both models, we decompose all parameters involved in our block of interest. We decompose those tensors into tensors ~1/10 of each of their original dimensions. For tiny-stories-8M this looks like:
\begin{verbatim}
transformer.h.4.attn.attention.k_proj.weight: [25, 25]
transformer.h.4.attn.attention.v_proj.weight: [25, 25]
transformer.h.4.attn.attention.q_proj.weight: [25, 25]
transformer.h.4.attn.attention.out_proj.weight: [25, 25]
transformer.h.4.attn.attention.out_proj.bias: [25]
\end{verbatim}

For mobilenet-v3-small this looks like:

\begin{verbatim}
features.7.block.0.0.weight: [12, 4, 1, 1]
features.7.block.0.1.weight: [12] 
features.7.block.0.1.bias: [12] 
features.7.block.1.0.weight: [12, 1, 5, 5]
features.7.block.1.1.weight: [12] 
features.7.block.1.1.bias: [12] 
features.7.block.2.fc1.weight: [3, 12, 1, 1] 
features.7.block.2.fc1.bias: [3]
features.7.block.2.fc2.weight: [12, 3, 1, 1]
features.7.block.2.fc2.bias: [12]
features.7.block.3.0.weight: [4, 12, 1, 1]
features.7.block.3.1.weight: [4]
features.7.block.3.1.bias: [4]
\end{verbatim}

\section{Supplemental Figures}

\begin{figure}[ht]
  \centerline{\includegraphics{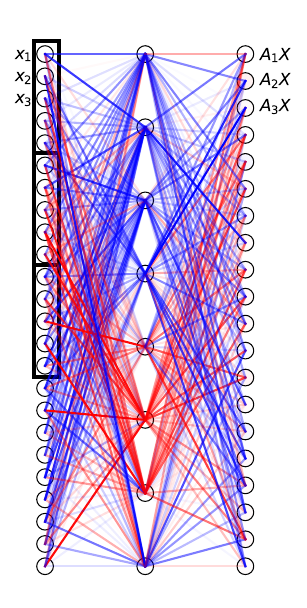}}
  \centering
  \caption{The full architecture of high rank circuit toy model (model C).}\label{fig:s1_high_rank_circuit_setup}
\end{figure}

\begin{figure}[ht]
  \centerline{\includegraphics{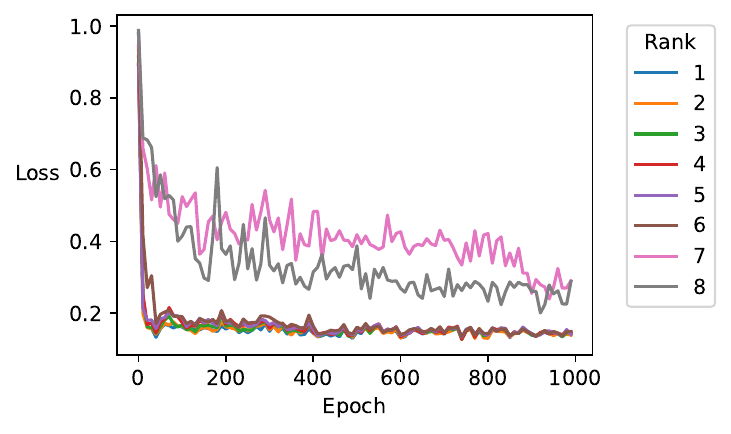}}
  \centering
  \caption{Reconstruction Loss vs Rank of the multi-feature/higher rank circuits.}\label{fig:s5_high_rank_circuits_loss_vs_rank}
\end{figure}
 
\begin{figure}[ht]
  \centering
  \caption{Decomposing the toy model of high rank circuits into different numbers of subnetworks.}\label{fig:s6_high_rank_decompositions}
  \begin{minipage}{\textwidth} 
  \centering
  \begin{tabular}{c} 
    \begin{subfigure}{.6\textwidth}
    \centering
    \includegraphics[width=\linewidth]{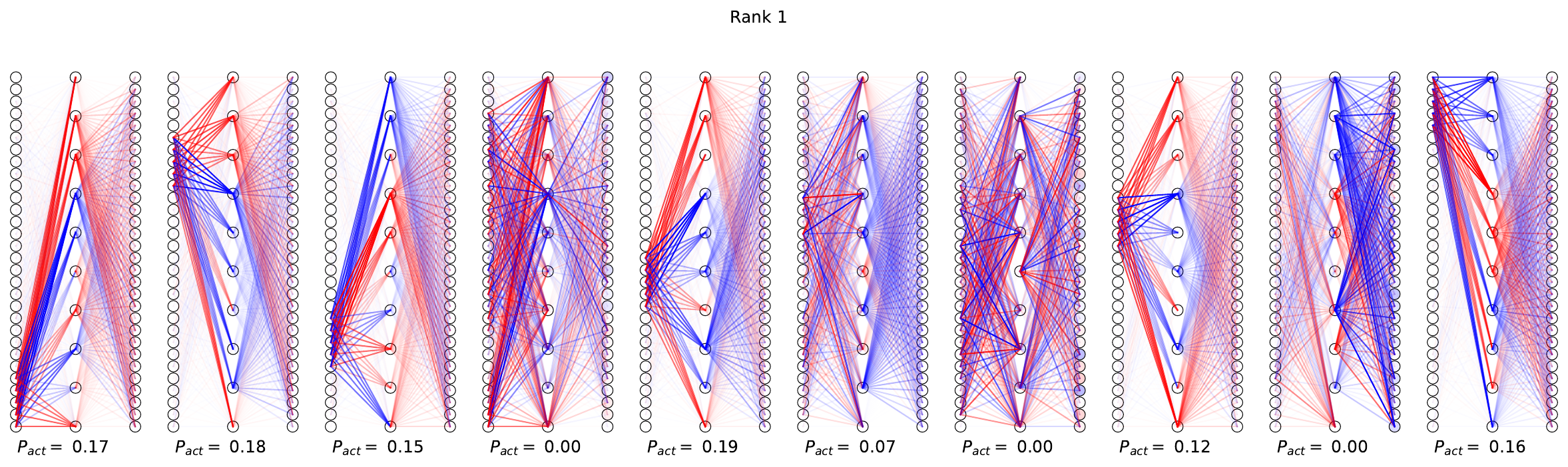}
    \caption{Rank-1 Networks}
    \end{subfigure} \\
    \begin{subfigure}{0.6\textwidth}
    \centering
    \includegraphics[width=\linewidth]{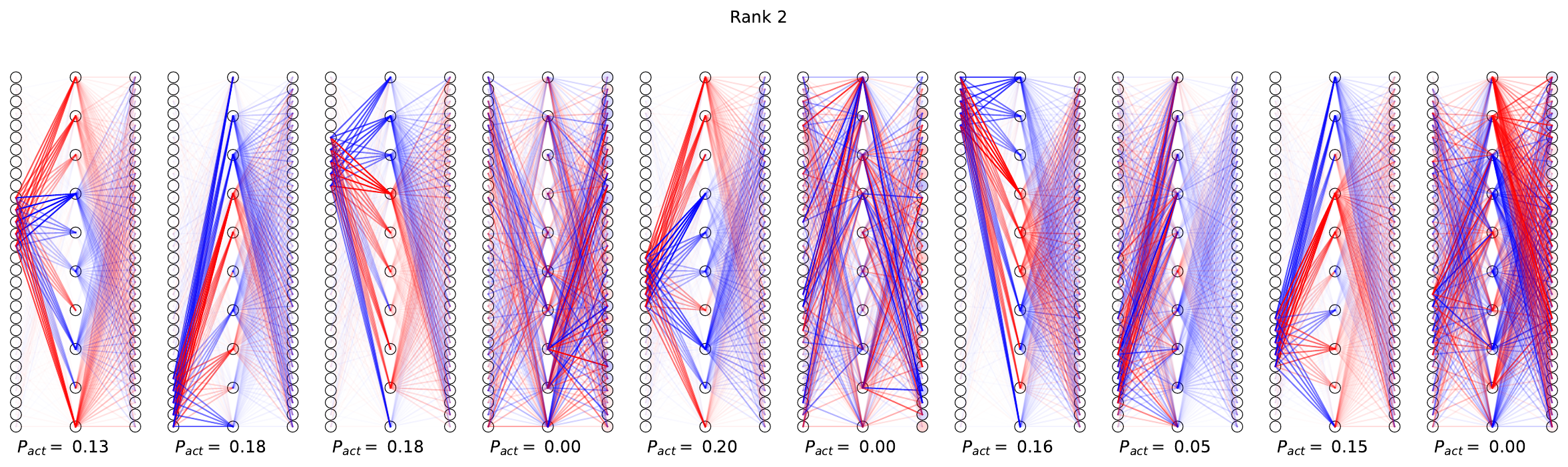}
    \caption{Rank-2 Networks}
    \end{subfigure} \\
    \begin{subfigure}{0.6\textwidth}
    \centering
    \includegraphics[width=\linewidth]{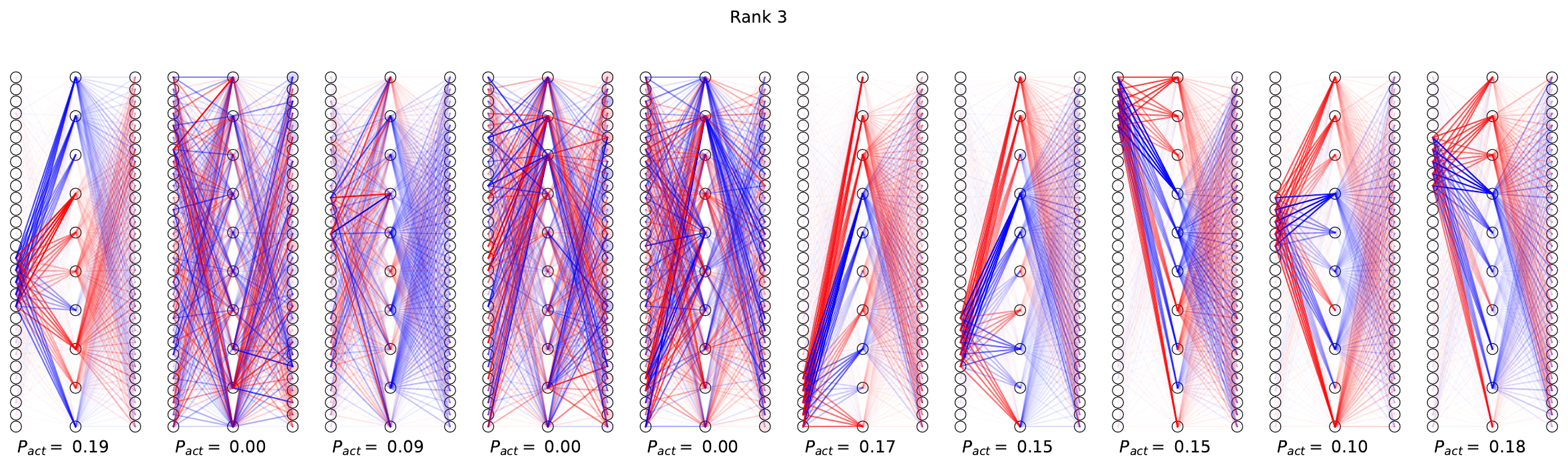}
    \caption{Rank-3 Networks}
    \end{subfigure} \\
    \begin{subfigure}{0.6\textwidth}
    \centering
    \includegraphics[width=\linewidth]{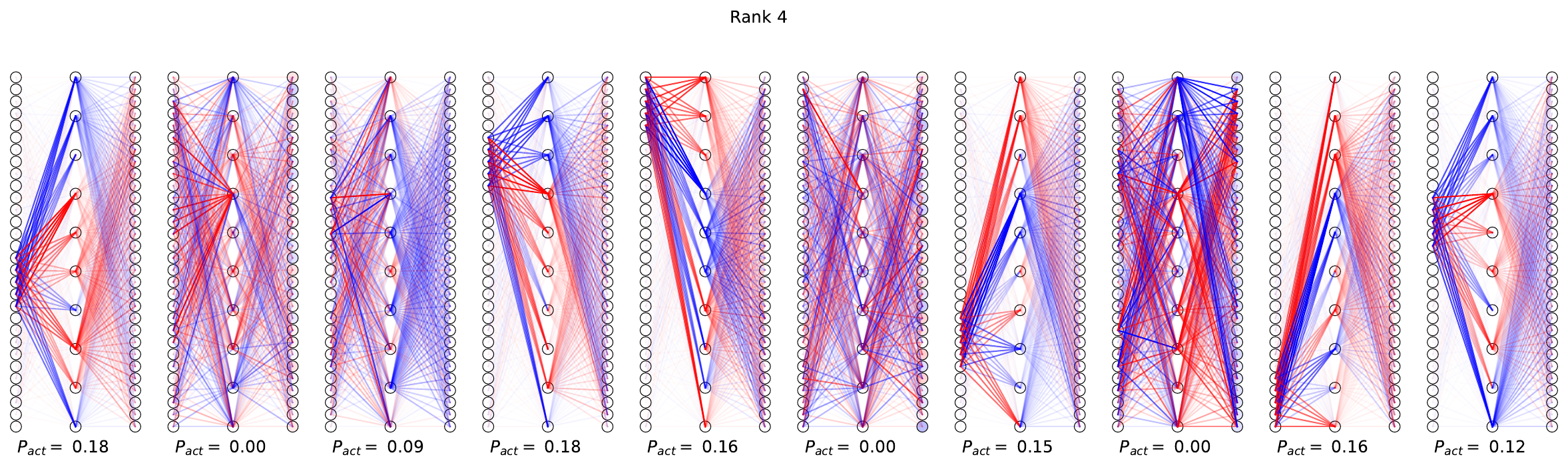}
    \caption{Rank-4 Networks}
    \end{subfigure} \\
    \begin{subfigure}{0.6\textwidth}
    \centering
    \includegraphics[width=\linewidth]{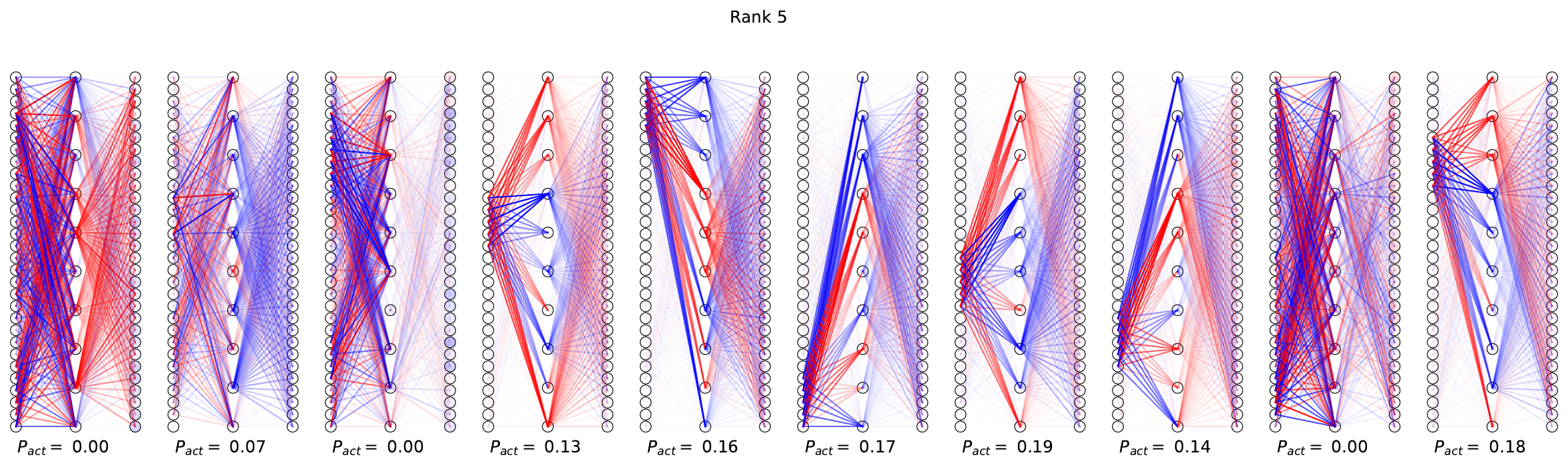}
    \caption{Rank-5 Networks}
    \end{subfigure} 
  \end{tabular}
  \end{minipage}

\end{figure}

\begin{figure}[ht]
  \centerline{\includegraphics[width=\textwidth]{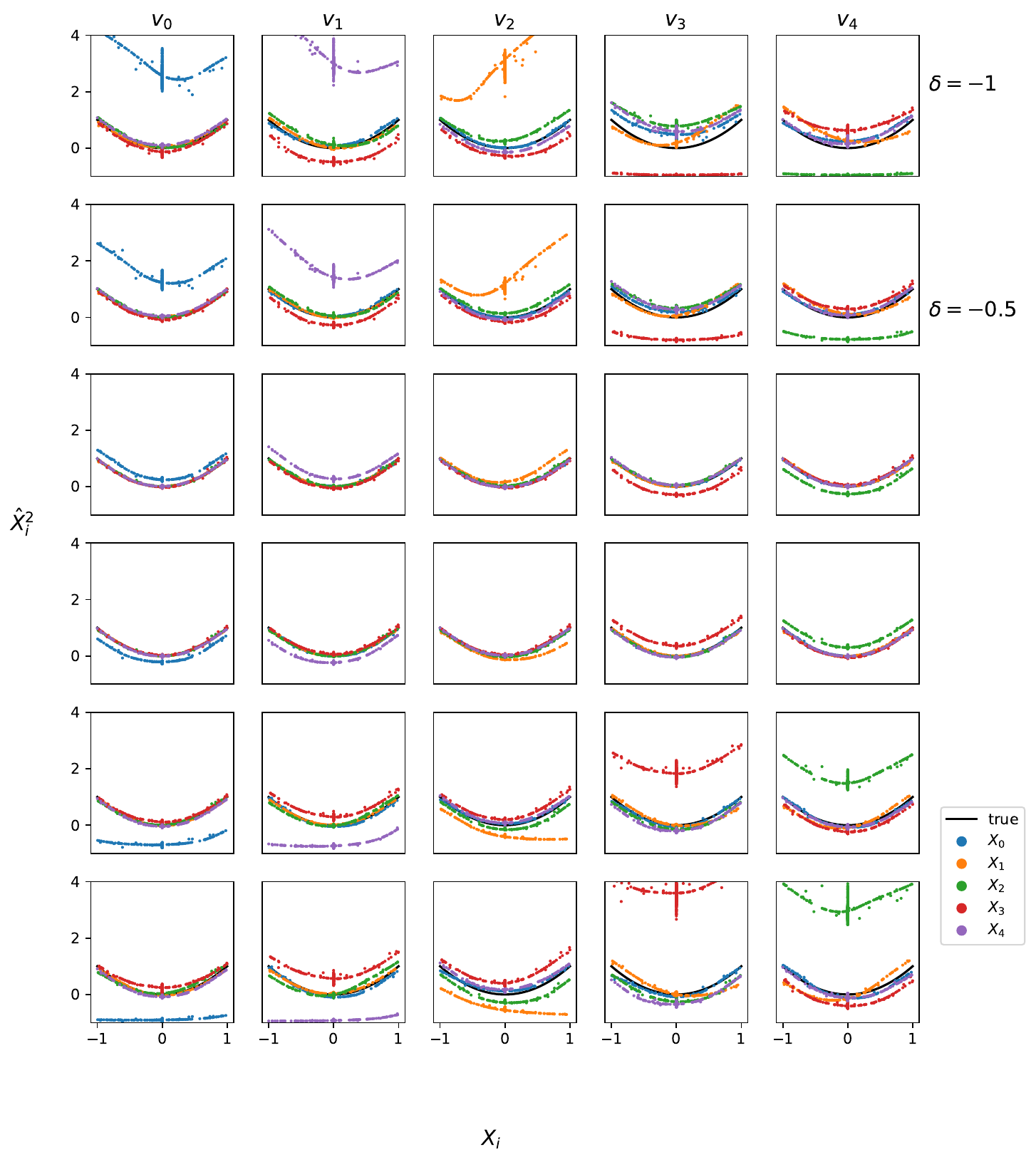}}
  \centering
  \caption{Effects of intervening on each of the subnetworks of the $X \mapsto X^2$ model.}\label{fig:s7_squared_intervention_more_deltas}
\end{figure}

\begin{figure}[ht]
  \centerline{\includegraphics[width=\textwidth]{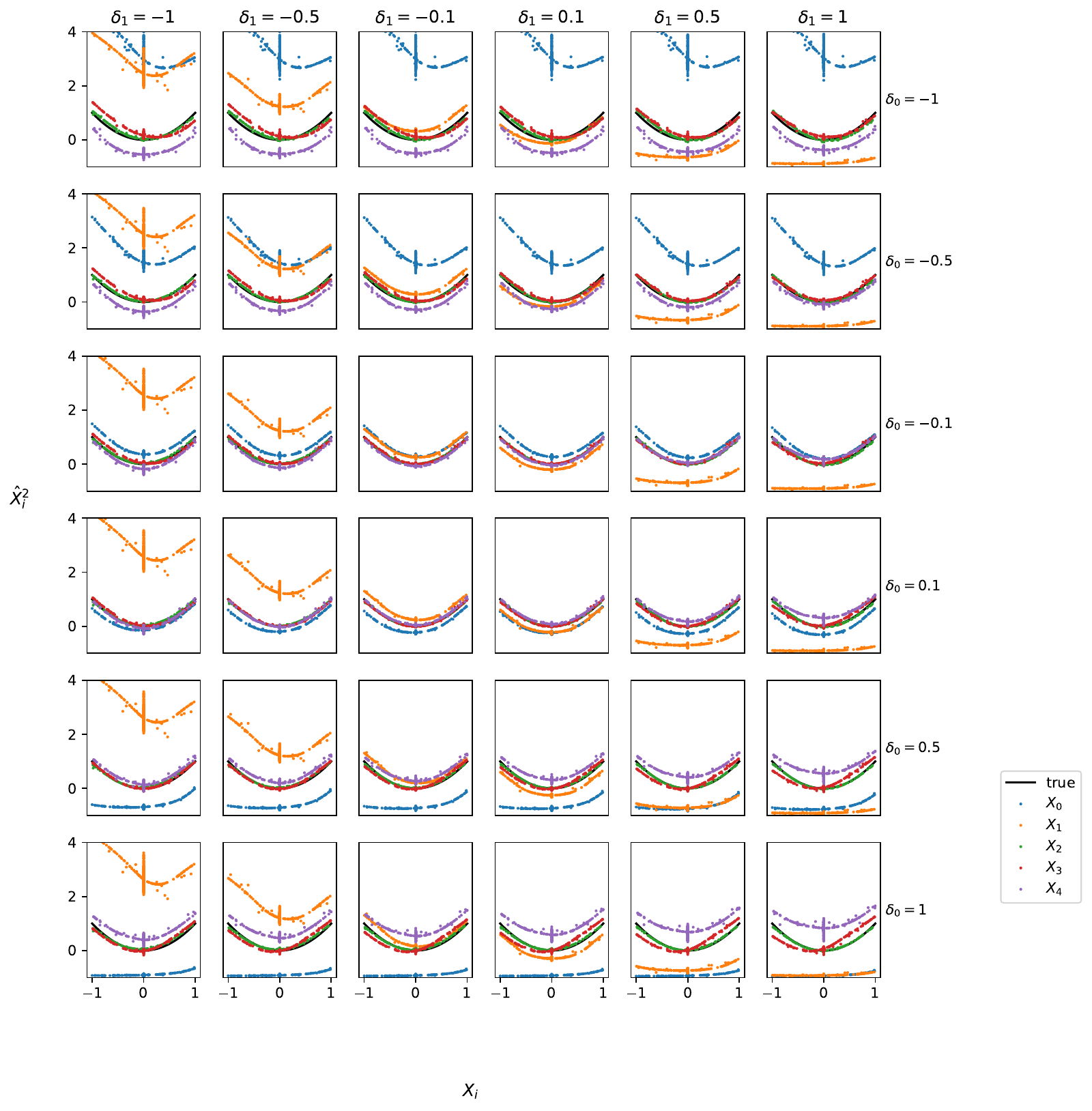}}
  \centering
  \caption{Effects of intervening with multiple subnetworks ($v_0$ on the x-axis, $v_1$ on the y-axis) at once.}\label{fig:s9_squared_intervention_multi_features}
\end{figure}



\begin{figure}[ht]
  \centering
  \caption{Decomposing the $X \mapsto X^2$ model into different numbers of subnetworks}\label{fig:s10_squared_decompositions_features}
  \begin{minipage}{\textwidth} 
  \centering
  \begin{tabular}{c} 
    \begin{subfigure}{\textwidth}
    \centering
    \includegraphics[width=.3\linewidth]{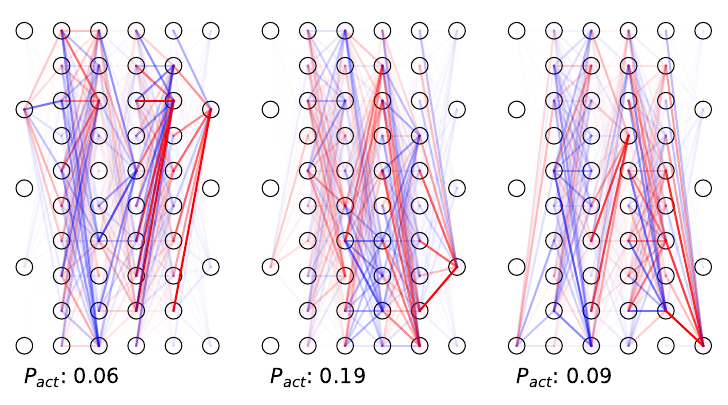}
    \caption{3 rank-1 Networks}
    \end{subfigure}\\
    \begin{subfigure}{\textwidth}
    \centering
    \includegraphics[width=.3\linewidth]{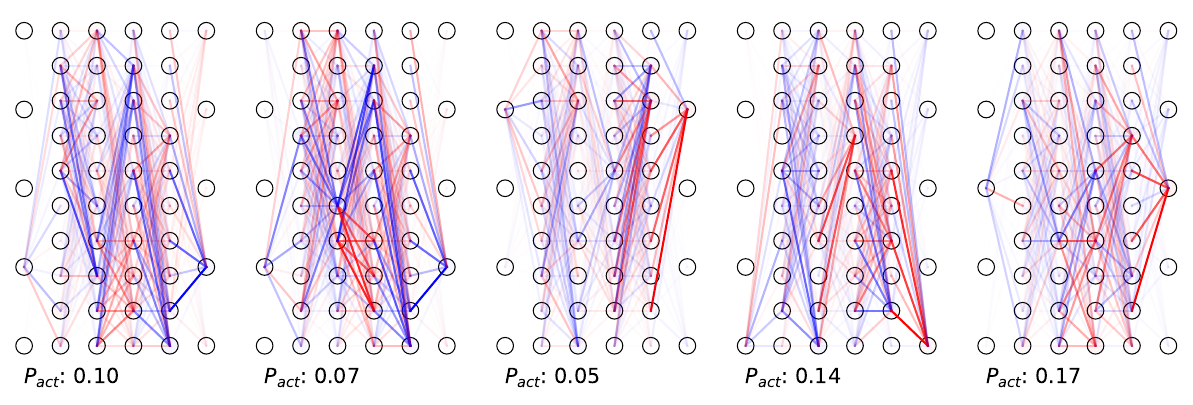}
    \caption{5 rank-1 Networks}
    \end{subfigure} \\ 
    \begin{subfigure}{\textwidth}
    \centering
    \includegraphics[width=\linewidth]{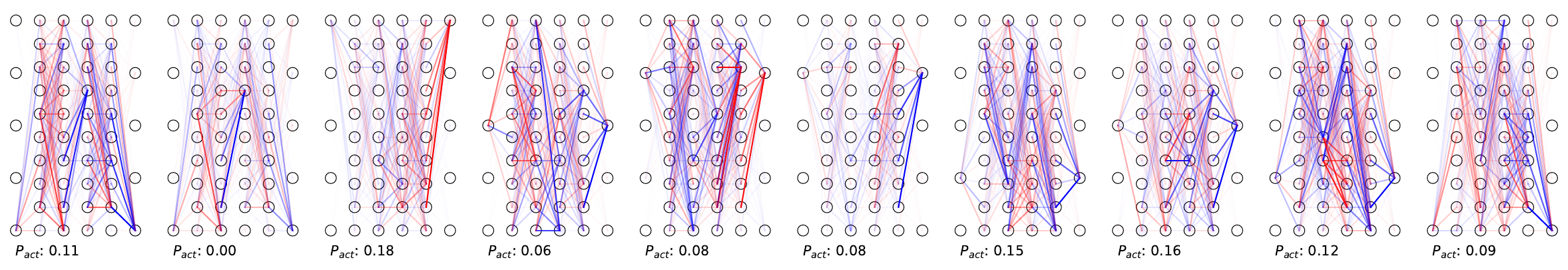}
    \caption{10 rank-1 Networks}
    \end{subfigure} \\ 
    \begin{subfigure}{\textwidth}
    \centering
    \includegraphics[width=\linewidth]{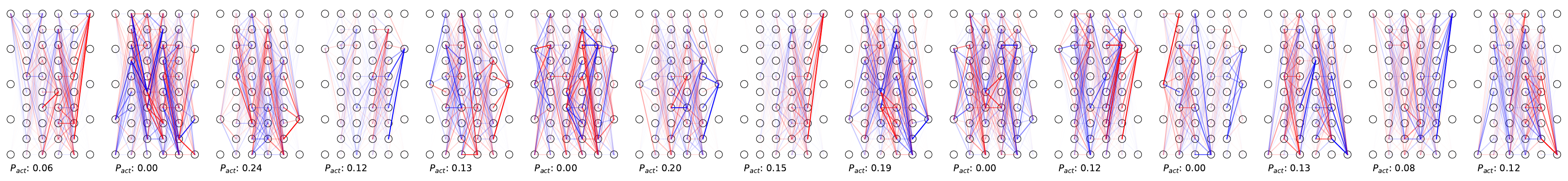}
    \caption{15 rank-1 Networks}
    \end{subfigure} 
  \end{tabular}
  \end{minipage}

\end{figure}

\begin{figure}[ht]
  \centerline{\includegraphics[width=\textwidth]{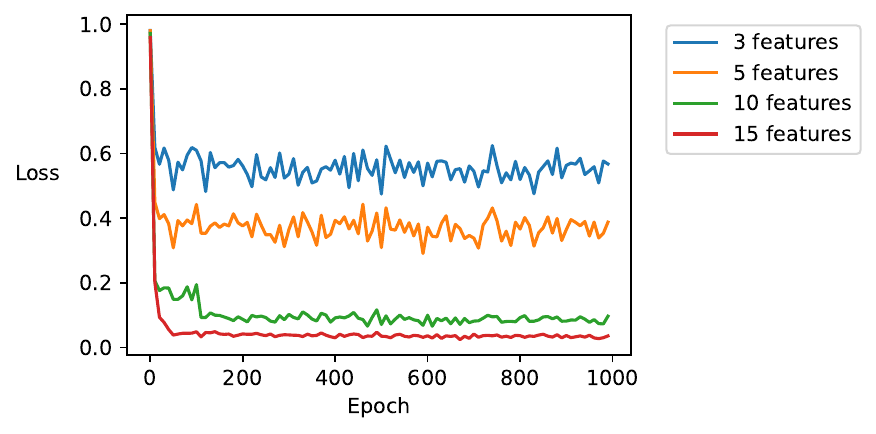}}
  \centering
  \caption{Training loss vs. number of subnetworks for the $X \mapsto X^2$ model}\label{fig:s11_squared_features_vs_loss}
\end{figure}

\onecolumn
\subsubsection{tiny-stories-8M Full Decomposition}\label{tab:transformer_full}
The full set of subnetworks (with $P_{act}>0$), most affected samples, and their most affected logits for the tiny-stories-8M decomposition. We list the subnetwork ID and $P_{act}$, show the most affected samples, and for each sample show the logits with the highest gradient with respect to the subnetwork.

\footnotesize
\begin{longtable}{|p{0.15\textwidth}|p{0.45\textwidth}|p{0.35\textwidth}|}
\hline
\textbf{Id ($P_{act}$)} & \textbf{Input Text} & \textbf{Top Logits} \\
\hline
\endfirsthead 
\hline
\textbf{Id ($P_{act}$)} & \textbf{Input Text} & \textbf{Top Logits} \\
\hline
\endhead

\hline
\multicolumn{3}{r}{\textit{Continued on next page}} \\
\hline
\endfoot

\hline
\endlastfoot
& & \\
\multirow{5}{*}{\textbf{0 (0.072)}} & be more careful when eating spicy food. From that & day, day, Monday, side, night \\
& too because she helped the bird. From that & day, side, umm, ts, Balls \\
& she should have been more careful. From that & day, day, side, cers, ts \\
& tummy hurt. From that & day, side, Balls, acas, ters \\
& . From that & day, side, acas, cers, Balls \\
& & \\
\multirow{5}{*}{\textbf{1 (0.114)}} & me." Lily smiles and claps her & hands, voices, mouths, faces, oves \\
& sit next to each & other, other, their, our, ours \\
& time, there was a little boy named Timmy. & Tim, One, They, It, There \\
& on a camping trip. Timmy was very excited!  As & they, their, them, They, theirs \\
& saw a cat on a tree. He wanted to be & friends, their, animals, our, together \\
& & \\
\multirow{5}{*}{\textbf{2 (0.036)}} & They are sad. They want to see the treasure. & <|endoftext|>, They, ", The, But \\
& car, a flower and a star. & <|endoftext|>, ", They, , The \\
& did not know why. & <|endoftext|>, The, ", , They \\
& and loud. They did not hear their mom calling them. & <|endoftext|>, ", , They, The \\
& basket and the knife behind. & <|endoftext|>, ", , They, The \\
& & \\
\multirow{5}{*}{\textbf{3 (0.063)}} & fun day at the park.Once upon a time, there was a boy named & Tim, Jack, James, Lily, Alex \\
& ."Once upon a time, there was a boy named & Tim, Jack, Lily, Ben, James \\
& They pretended to be kings and & queens, she, princes, her, She \\
& in the future.Once upon a time, there was a big elephant named & Ellie, Ell, Daisy, Grace, Lily \\
& Once upon a time, there was a boy named Tim. & He, Every, She, They, Sue \\
& & \\
\multirow{5}{*}{\textbf{4 (0.111)}} & , doctor. Thank you, mom. Thank you & ,, Star, ider, printers, Auto \\
& . It looked happy and friendly. "See, Lily and Ben & ,, ?", !", ?!", !, \\
& panic and cry. Her mom knew it & and, ., ,, would, wasn \\
& , Mom. Please, can & we, I, pie, soldiers, Hood \\
& 't worry, Timmy & ., ,", !, .", !" \\
& & \\
\multirow{5}{*}{\textbf{5 (0.107)}} & together.Once upon & a, an, SEC, irled, clip \\
& best friends.Once upon & a, an, SEC, clip, irled \\
& , so they stay colorful and clean."Once upon & a, an, orse, ship, ream \\
& you for being so persistent, daddy."Once upon & a, an, ud, orse, SEC \\
& became good friends.Once upon & a, an, clip, SEC, irled \\
& & \\
\multirow{5}{*}{\textbf{6 (0.229)}} & to play. They went on the swings and the slide. Lily had so & much, killing, backdrop, doorstep, ocus \\
& way there, he got lost. He couldn't find his & way, results, rr, umbers, For \\
& noises. Tom had a small car that could go fast and beep. Lily & wanted, liked, was, loved, and \\
& teddy bear and had a lot of & fun, adventure, lots, daring, thrilling \\
& jump in. They had so & much, satisfying, wr, Ah, izz \\
& & \\
\multirow{5}{*}{\textbf{7 (0.085)}} & They made a new friend. They were very happy & ., elegance, effective, ulent, val \\
& your dragon." They all laughed and hugged. They were happy and glad & ., unky, utch, aved, cog \\
& ice cream. It was cold and sweet. They were very happy & ., iot, error, angled, uld \\
& They are sad & and, stack, Figure, rast, wered \\
& and milk. Lily was very happy & and, ., redible, ulent, arise \\
& & \\
\multirow{5}{*}{\textbf{8 (0.067)}} & said to the plant, "We are sorry, plant. We did & not, t, opposite, roll, pe \\
& mister," Tom says. "We did & not, t, ts, still, steadily \\
& worm," she said. "I'm sorry, mushroom. I did & not, t, roll, ts, sly \\
& marks on the wall too, but Mommy does & not, .), trade, ll, fir \\
& The bee was on the apple. It was angry and scared. It did & not, roll, ves, dig, not \\
& & \\
\multirow{5}{*}{\textbf{9 (0.177)}} & ran to hide behind a tree. She peek & ed, apped, red, faced, Buddha \\
& it in the lock. They push and pull, but nothing & happened, comes, works, is, breaks \\
& Mommy will be angry. He says, & ", ulating, ver, attered, atter \\
& you for the treat!" Spot bark & ed, ked, red, apped, led \\
& . Lily pet & ted, ged, led, aced, oled \\
& & \\
\multirow{5}{*}{\textbf{10 (0.072)}} & that day on & ., the, she, he, that \\
& that day on & ., the, she, he, that \\
& that day on & ., the, she, he, that \\
& a little girl named Lily. She loved to play in the park with her friends & ., .), questions, app, ongs \\
& named Lily. She loved to play outside in the sun with her friends & ., tricks, .', questions, cycle \\
& & \\
\multirow{5}{*}{\textbf{11 (0.062)}} & at the shell. They looked at their mom. They looked at each other. & They, bits, pper, circuits, uff \\
& floor. They are sorry. They do not want to make mom sad. & They, rily, acks, spotlight, bits \\
& . Lily and Ben look at each other. They are scared. & They, acks, over, bits, laz \\
& tall man comes to the tree. He has a hat and a coat. & He, acks, ogged, itting, ung \\
& !" Anna does not listen to Ben. She thinks he is silly. & She, ito, ails, ative, acquired \\
& & \\
\multirow{5}{*}{\textbf{12 (0.063)}} & you," said Finny sadly. "Don't worry, Fin & ny, ster, ur, Duck, armed \\
& named Sue. She had a big, red ball. " & Let, Ball, Hi, Can, Wow \\
& "Ben, Ben, grab the stick!" she shouted. " & Give, You, The, That, This \\
& football. "Wow, look at this football!" Ben says. " & It, We, It, it, Mine \\
& not listen to Mia. He wanted to win. He did & not, disagree, surrender, yoga, being \\
& & \\
\multirow{5}{*}{\textbf{13 (0.031)}} & you want some?" & L, Anna, Tim, S, M \\
& . They hugged Mom and Dad. & The, L, M, S, Then \\
& say sorry. & L, S, She, Anna, He \\
& scared too. & The, One, L, She, He \\
& & One, L, The, When, As \\
& & \\
\multirow{5}{*}{\textbf{14 (0.191)}} & every day. The plant had green & plants, roots, needles, stems, and \\
& together in the leaves. The end.<|endoftext|> & ,  , , A, From \\
& playing, he saw a big hole in the & garden, wall, fence, middle, backyard \\
& on walks and helped other & children, people, kids, creatures, young \\
& but there was none. The sun was getting hotter and the goat was getting thirst & ., and, der, y, of \\
& & \\
\multirow{5}{*}{\textbf{15 (0.178)}} & . Ducky tri & ump, pping, onto, over, inged \\
& hill very fast. Tim and Sue laughed and cl & apping, ap, aps, amb, ink \\
& ogged and played, having a lot of fun. As they j & ogging, olly, ogg, umbled, ogs \\
& They both pulled and tug & on, ging, ., and, ed \\
& named Max & ., playing, coming, looking, walking \\
& & \\
\multirow{5}{*}{\textbf{16 (0.028)}} & it first!" Sara says. "We want to see the treasure!" & Ben, Tom, She, she, Tim \\
& . They are not ours to take. They are the sea's to give." & They, Tom, Ben, Mom, Tim \\
& race!" Ben said. "I bet I can go faster than you!" & He, Lily, Mia, , he \\
& is not good to touch. Mom said some mushrooms are bad." & But, Mom, They, Ben, Lily \\
& chicken too. They are all good for you." & They, Mom, , The, Lily \\
& & \\
\multirow{5}{*}{\textbf{17 (0.152)}} & the park with their bikes. They liked to ride fast and make noises. They & saw, heard, met, played, ate \\
& up the ball with its be & ak, ams, ck, umb, arrow \\
& You wasted a lot of food and drinks. You have & to, disturbed, wandered, shown, bumped \\
& , you can," Lily and Tom said, nodding. "But you have & to, disturbed, delayed, pulled, forced \\
& eat avocados, they were her & best, friends, new, very, special \\
& & \\
\multirow{5}{*}{\textbf{18 (0.060)}} & . It was your treasure." Ben shook his & head, izing, Warning, iated, alking \\
& . Lily and Ben look at each & other, enlarged, OUT, pping, heit \\
& at the shell. They looked at their mom. They looked at each & other, wait, pace, lower, bribe \\
& clumsy, Sam," Tom said, shaking his & head, neck, chin, heads, eyebrows \\
& chicken too. They are all good for you." Tom shook his & head, Warning, FUN, izing, Save \\
& & \\
\multirow{5}{*}{\textbf{19 (0.060)}} & !" Anna does & not, blush, Justin, Alan, Harry \\
& . Sara did & not, blush, word, ordering, waitress \\
& , you can," Lily and Tom said, nodding. "But you have & to, aker, ican, ator, eting \\
& want to play anymore. This is too difficult." But Lily did & not, if, ake, girl, Should \\
& ask God to help you eat your soup." Tom did & not, word, blush, asking, orman \\
& & \\
\multirow{5}{*}{\textbf{20 (0.105)}} & . It grew new leaves and flowers. Anna and Ben were & amazed, excited, sad, curious, not \\
& animals like lions and monkeys. It was so much & bigger, better, that, more, hard \\
& jump in. They had so much & energy, to, that, stuff, time \\
& dolls. Lily was & a, not, playing, the, excited \\
& bird on a branch. The bird was & blue, sitting, singing, yellow, flying \\
& & \\
\multirow{5}{*}{\textbf{21 (0.094)}} & dad were hurt too. They went to the & hospital, doctor, nurse, car, pool \\
& They hide the letter under the & couch, bed, sofa, table, slide \\
& They could play on the & swings, beach, subway, climbers, Safari \\
& the old lady talked on the & phone, telephone, cellphone, plaza, cafeteria \\
& to see who could get the best score. Tim threw the & ball, balls, basketball, trash, seeds \\
& & \\
\multirow{5}{*}{\textbf{22 (0.090)}} & a little & bit, bird, scared, while, too \\
& a little & bit, bird, scared, while, too \\
& , a little & bird, mouse, dog, bug, red \\
& a yummy & food, soup, and, ,, dinner \\
& a little & bit, bird, scared, while, too \\
& & \\
\multirow{5}{*}{\textbf{23 (0.066)}} & clumsy, Sam," Tom said, shaking his & head, tail, ow, spine, ows \\
& loved to play with his toy & car, animals, gun, boat, truck \\
& playing, he saw a big hole in the & fence, wall, tree, garden, corner \\
& my. Timmy loved to play with his toy & car, gun, hammer, je, boat \\
& grabbed her cray & on, in, ane, ions, ip \\
& & \\
\multirow{5}{*}{\textbf{24 (0.063)}} & The end.Once upon a time, there was a little girl named Lily. & She, He, Max, Emma, Tim \\
& upon a time, there was a little girl named Lily. & She, He, Max, Tim, Tom \\
& .Once upon a time, there was a little girl named Lily. & She, He, Max, Tim, Emma \\
& upon a time, there was a little girl named Lily. & She, He, Max, Tim, Tom \\
& upon a time, there was a little girl named Lily. & She, He, Max, Tim, Tom \\
& & \\
\multirow{5}{*}{\textbf{25 (0.094)}} & .Anna liked to examine things. She liked & to, touching, explore, smoot, pile \\
& ! Thank you so & long, !", high, hard, fast \\
& to go home. His friend asked him what was & going, in, happening, inside, he \\
& his arm. His mom took him to the & hospital, park, store, bathroom, nurse \\
& fun that she didn't & want, ., mind, see, notice \\
& & \\
\multirow{5}{*}{\textbf{26 (0.199)}} & to investigate and found shiny rocks that spark & ly, ened, les, le, bled \\
& garden. He sne & aked, wed, uned, amped, ound \\
& shell back. She tried to grab it from Tom's hand. " & No, Mine, O, Me, Please \\
& Lily. She had a cup that she loved to drink juice from every & day, time, week, evening, time \\
& Emma heard her sister's scream and asked, " & Is, Why, Are, Please, Who \\
& & \\
\multirow{5}{*}{\textbf{27 (0.125)}} & ." They ran back & home, and, ,, together, . \\
& ran to hide behind a tree. She peeked & behind, around, her, at, inside \\
& ily wanted to touch it anyway. She reached & for, her, the, it, his \\
& . He picked it & and, carefully, off, with, out \\
& became good friends.Once upon a time, a bird wanted to fly high & and, ., ,, like, above \\
& & \\
\multirow{5}{*}{\textbf{28 (0.065)}} & found you!" It was her friend, Tim. & He, ", They, She, Tim \\
& on a camping trip. Timmy was very excited! & He, He, was, to, They \\
& outside. & He, She,  , , The \\
& , but they were too messy. & They, The,  , Suddenly, Tim \\
& to climb on. & He, She, The,  , \\
& & \\
\multirow{5}{*}{\textbf{29 (0.030)}} & told her not to worry and that she would take care & ., for, and, about, when \\
& a big house with a lot & to, us, ., er, more \\
& even higher!Once upon a time, there was a big, strong robot made & ., out, up, from, - \\
& played in the garden and took care & to, for, ., with, every \\
& to play and run all day. One day, Tim found a big bag & ., in, and, on, with \\
& & \\
\multirow{5}{*}{\textbf{30 (0.048)}} & She did not see her & ., and, feet, hand, Mom \\
& in the bathtub. She did not hear her & Mom, voice, mother, big, brother \\
& She said to her & ,, daughter, little, friend, Mom \\
& outside.  Lily told her & mom, ,, grandma, Mom, that \\
& night. One day, she told her & friend, friends, Mom, parents, mother \\
& & \\
\multirow{5}{*}{\textbf{31 (0.062)}} & . She smiles and says, & ", attered, ayed, atter, appers \\
& Lily nodded and said, & ", attered, cher, ayed, atter \\
& Mia hugged Ben and said, & ", ayed, atter, attered, havoc \\
& . She gives each doll a cup and a plate. She says, & ", attered, led, ayed, umbled \\
& happy to see Anna's spoon. They say, & ", attered, atter, ored, ico \\
& & \\
\multirow{5}{*}{\textbf{32 (0.083)}} & . Lily wanted to join in on the fun, but her mom told & them, she, the, Lily, it \\
& earlier, but he still wanted to help her. He went over and helped & the, his, Lily, pick, them \\
& very happy. Tim's mom was proud of & Tim, his, them, her, the \\
& Mom smiled and hugged them. She gave & the, her, Lily, their, back \\
& day, Lily's mom asked & the, if, him, Lily, them \\
& & \\
\multirow{5}{*}{\textbf{33 (0.044)}} & be thoughtful and careful when helping others.Once upon & the, time, , first, finding \\
& , Monkey always kept his room tidy just like Ellie's.Once upon & the, time, playing, Lily, \\
& . The end.Once upon & the, time, first, then, , \\
& and making more pictures together.Once upon & the, time, , first, an \\
& with his dad and ride his bike with gears on the clear path.Once upon & the, time, first, to, \\
& & \\
\multirow{5}{*}{\textbf{34 (0.040)}} & & Once, The, One, L, Tom \\
& & Once, The, One, L, Tom \\
& & Once, The, One, L, Tom \\
& & Once, The, One, L, Tom \\
& & Once, The, One, L, Tom \\
& & \\
\multirow{5}{*}{\textbf{35 (0.070)}} & spider was about & to, beverages, rene, agons, Spears \\
& still sounded bad. He was about & to, offerings, agons, rig, unky \\
& . He didn't mean & to, fullest, custom, destination, idol \\
& want to go to the police. They decide & to, conclusions, erer, fascination, prod \\
& careful not & to, iot, plaza, aned, continents \\
& & \\
\multirow{5}{*}{\textbf{36 (0.091)}} & -cream, and had lots of fun at the park. The end & !, ,, of, .", \\
& are gone. The end & .", is, !", of, result \\
& . The end & .", !, was, of, result \\
& . The end & .", !, was, of, result \\
& the flag wave in the wind. The end & !, was, of, .", , \\
& & \\
\multirow{5}{*}{\textbf{37 (0.070)}} & !" Lily said. "Yes, it is," her & mom, aining, ably, irs, irted \\
& my didn't want to share his toys, so his & mom, inges, aining, IN, irs \\
& fun that she didn't want to leave. But her & mom, aining, lier, iment, inges \\
& cereal for breakfast every day. One day, her & mom, lier, irs, irting, piece \\
& After they finished playing, Timmy went home. Lily's & mom, lier, aining, purposes, arl \\
& & \\
\multirow{5}{*}{\textbf{38 (0.097)}} & were packing, Timmy's mom reminded him to bring his flashlight. & She, They, He, But, \\
& say they did not open the box. & <|endoftext|>, .", But, They, . \\
& She loved to walk on the trail with her dog, Max. & They, Max, , The, She \\
& back. & <|endoftext|>, ,", .", They, too \\
& bear. They tell them that they have to wait for Christmas. & <|endoftext|>, , <, ., They \\
& & \\
\multirow{5}{*}{\textbf{39 (0.079)}} & didn't & know, want, like, understand, think \\
& didn't & know, want, like, understand, think \\
& didn't & know, want, like, understand, think \\
& sad and didn't & know, want, understand, care, quit \\
& does not & like, know, want, hear, understand \\
& & \\
\multirow{5}{*}{\textbf{40 (0.192)}} & the rock! Lily was upset and scared. She & didn, really, questioned, rew, Wow \\
& he was very sad. Lily & wanted, asked, didn, told, said \\
& But we found them here," Ben & says, said, insisted, suggested, wiped \\
& , scary fox came into the garden. Bongo & didn, was, felt, wanted, did \\
& had passed away. Lily & was, felt, didn, went, missed \\
& & \\
\multirow{5}{*}{\textbf{41 (0.118)}} & and reached for an apple. But she did not & see, wind, Wait, trips, trip \\
& !" Sara and Ben are scared. They do not know & what, where, moms, sure, shore \\
& untied! Timmy didn't & know, hesitate, hate, doubt, 've \\
& sad and didn't & know, knowing, wanting, being, noticing \\
& were stuck. Lily started to feel scared and silly. She didn't & know, knowing, wanting, extra, being \\
& & \\
\multirow{5}{*}{\textbf{42 (0.113)}} & with his ball. One day, & Tim, Benny, Max, Twe, Remy \\
& restless.  As Timmy rode his bike, & he, unison, aining, ainer, centers \\
& One day, & she, Lily, Tim, Benny, Max \\
& wet. One day, & Lily, she, Tim, Max, Benny \\
& Timmy. One day, & Tim, Nem, Remy, Nut, T \\
& & \\
\multirow{5}{*}{\textbf{43 (0.009)}} & fun day at the park.Once upon a time, there & was, extingu, ixtures, manship, burden \\
& that might have something yummy inside.Once upon a time, there & was, manship, Shadow, defense, yles \\
& a time, there & was, ixtures, accurate, yles, manship \\
& truck all day long.Once upon a time, there & was, ixtures, manship, yles, backdrop \\
& them disappear again.Once upon a time, there & was, manship, yles, tripod, ixtures \\
& & \\
\multirow{5}{*}{\textbf{44 (0.085)}} & dough. She put the cookies in & the, her, my, Becky, Mrs \\
& to play games with his friends in & the, Christ, elled, ussed, aming \\
& He loved to play with his ball in & the, The, Lyn, His, Ray \\
& play and run in & the, sect, Christ, oned, elled \\
& men and playing in & the, Lyn, Christ, Den, rod \\
& & \\
\multirow{5}{*}{\textbf{45 (0.035)}} & Once upon a time, there was & an, the, one, two, something \\
& Once upon a time, there was & an, the, one, two, something \\
& Once upon a time, there was & an, the, one, two, something \\
& they lived happily ever after.Once upon a time, there was & an, the, one, another, Lily \\
& ."Once upon a time, there was & an, the, Lily, one, something \\
& & \\
\multirow{5}{*}{\textbf{46 (0.061)}} & his shoe. Timmy was so & excited, proud, sad, surprised, embarrassed \\
& mom looked around and found it under the bed. Timmy was so & excited, proud, surprised, glad, grateful \\
& my was so & excited, proud, sad, surprised, scared \\
& was and decided to permit him to play with his skull again. Spot was so & excited, grateful, glad, proud, surprised \\
& thank you. Lily was so & excited, glad, proud, grateful, sad \\
& & \\
\multirow{5}{*}{\textbf{47 (0.075)}} & you want." Tim said, & ", ayer, est, ime, over \\
& Sue asked. Tim said, & ", apper, attered, ime, appers \\
& nice. Tom said, & ", attered, apper, ilt, iner \\
& faucet for the kitchen sink. Mia's mom said, & ", attered, appers, apper, umbled \\
& are you sad, Tom?" Tom replied, & ", "", anes, overs, ooters \\
& & \\
\multirow{5}{*}{\textbf{48 (0.357)}} & up the ball with its beak and brings & him, the, her, back, them \\
& to play with cars and balls and blocks. They go to the & park, same, zoo, library, beach \\
& It was yellow and black and very pretty. She ran & around, after, outside, and, inside \\
& saw a big dog running & around, in, across, after, up \\
& watch where he was going and tri & pping, ump, umble, umbles, ey \\
& & \\
\multirow{5}{*}{\textbf{49 (0.193)}} & went to the park with her mom and saw her friends playing hide-and- & and, go, tag, ider, pack \\
& her mommy said, trying to calm her & down, concentrate, responsibility, downstairs, concentration \\
& . Lily and Ben look at each & another, one, thing, ., toy \\
& was too heavy and slow. The bunny got away and the all & ion, that, the, bunny, of \\
& on, they went for walks in the park together and became good & at, and, -, players, siblings \\
& & \\
\multirow{5}{*}{\textbf{50 (0.082)}} & . Tim saw his friend, a big dog & ,, ., and, with, called \\
& with a smile.Once upon a time, there was a little girl & who, ., called, and, with \\
& .Once upon a time, there was a graceful cat & ., who, called, and, with \\
& see the beautiful yellow sunrise.Once upon a time, there was a boy & who, and, ., called, with \\
& .Once upon a time, there was a little girl & who, ., called, and, with \\
& & \\
\multirow{5}{*}{\textbf{51 (0.051)}} & my and daddy. One day, while swimming, Tim & my, cases, astic, certainty, Noise \\
& at the campsite, Tim & my, Christ, ISON, arten, Mood \\
& he accidentally bumped into the barrel and it started rolling. Tim & my, itate, generations, judgments, Staff \\
& 's legs got tired and they stopped to take a break. Tim & my, generations, adversary, itate, Long \\
& my. Tim & my, Forest, itate, oids, Christ \\
& & \\
\multirow{5}{*}{\textbf{52 (0.078)}} & basket and the knife behind. Dad did & not, warn, scare, poop, rier \\
& me." Lily smiles and cl & ucks, apped, ink, ags, ums \\
& to hurt you. Please forgive us." The plant did & not, Not, prick, Woo, scare \\
& had a black cat named Mittens. Mittens was very soft and c & uffy, agged, aged, led, owed \\
& She sees the letter. It is torn. She sigh & ., es, and, ing, again \\
& & \\
\multirow{5}{*}{\textbf{53 (0.101)}} & . Buzzy flew down and said, " & Hello, Hi, Thank, hello, Wow \\
& it to Ben. Ben kicks it back to Tom. They have & fun, ritz, rer, Absolutely, ream \\
& band-aid on it. He gives Lily a sticker and a l & ily, icks, olly, icked, kin \\
& it was time to go home. Timmy went to bed that & night, afternoon, game, chance, Friday \\
& shell back. She tried to grab it from Tom's hand. " & Give, Hey, Go, O, Come \\
& & \\
\multirow{5}{*}{\textbf{54 (0.114)}} & at first, but he decided to try it. Nemo and Crab & by, bles, iny, bly, as \\
& One day, she decided to examine the bath & tub, robe, ro, tub, bath \\
& her room. She puts the teaspoon in Anna & and, ., , ,, ' \\
& me." Lily smiles and cl & apped, ucks, ums, s, ink \\
& young boy named Tim found a dull, round rock. He picked it & and, out, from, with, , \\
& & \\
\multirow{5}{*}{\textbf{55 (0.048)}} & found you!" It was her friend, Tim. & He, ", They, Tim, It \\
& to climb on. & He, She, The, , \\
& and showed it to her dog. & She, , It, The, They \\
& was light, so Tim could pull it easily. & He, , The, Tim, They \\
& had touched the flower. & She, He,  , The, It \\
& & \\
\multirow{5}{*}{\textbf{56 (0.085)}} & dough. She put the cookies in & the, a, sacks, Lisa, Sue \\
& He saw her on & the, his, their, Wednesday, your \\
& back. Then he sees a duck. The duck is swimming in & a, an, nature, another, rivers \\
& their toys in & their, the, different, another, Mia \\
& . One day, she saw a butterfly flying in & a, the, her, an, nature \\
& & \\
\multirow{5}{*}{\textbf{57 (0.077)}} & , Lily wanted to try to lift a heavy frame all by & himself, itself, themselves, yourself, myself \\
& dog stopped barking and Timmy felt much better. He got & up, dry, bedroom, soak, mood \\
& Mom. They saw a big pond with many ducks and sw & am, an, immers, ucky, ishes \\
& and went outside to eat by & the, itself, his, its, her \\
& Max saw a big plane flying in the sky. Max barked excited & ly, eyes, p, en, bly \\
& & \\
\multirow{5}{*}{\textbf{58 (0.112)}} & left and right. They loved marching together. & , The, One, , They \\
& very nice, Mom," End said. & , ", , The, M \\
& be careful with fragile things. & , One, The, ", \\
& Max and they both had a great time chewing on it together. The & moral, little, sun, two, next \\
& mom was proud of her for being kind and sharing. & , , The, From, But \\
& & \\
\multirow{5}{*}{\textbf{59 (0.023)}} & to sleep." Tom gave back the jewelry and said, "Thank & you, background, ptions, mats, react \\
& Lily nodded and said, "Thank & you, opes, ptions, mats, speakers \\
& , "Thank & you, ptions, background, technique, bolts \\
& It looked happy. "Thank & you, ptions, opes, bolts, zel \\
& Ben smiled and said, "Thank & you, ptions, opes, background, bolts \\
& & \\
\multirow{5}{*}{\textbf{60 (0.053)}} & corn move back and & forth, appers, unfairly, apper, EST \\
& said she didn't know. Lily looked everywhere for her cup, & the, even, her, under, which \\
& ? I told you about the cable. You were not wise. & I, Next, Now, Do, How \\
& to read before she went to bed. Mia looked at the bookshe & read, books, book, was, r \\
& treasure. He hit the ice harder and & harder, faster, slower, easier, farther \\
& & \\
\multirow{5}{*}{\textbf{61 (0.151)}} & too because she helped the bird. From that & moment, night, ,, time, morning \\
& sharing all of their toy tools. From that & moment, time, ,, afternoon, night \\
& forgot about her knee. From that & moment, night, morning, time, afternoon \\
& up on the fridge. From that & night, moment, morning, ,, time \\
& and finally, they found the belt under Tom's bed. Tom was & happy, very, not, surprised, sad \\
& & \\
\multirow{5}{*}{\textbf{62 (0.029)}} & . She says, " & I, Thank, You, Don, Wow \\
& . He ate his celery. He was happy. He said, " & Thank, You, Wow, Pot, Work \\
& hugged Lily. " & I, Thank, It, You, Wow \\
& They hug mom. They say together. " & Thank, We, Can, I, You \\
& Mia hugged Ben and said, " & Thank, You, Don, Wow, Are \\
& & \\
\multirow{5}{*}{\textbf{63 (0.057)}} & had their wand and their bubbles. They did & not, ann, ales, pered, Lumin \\
& had to pick some onions for dinner. Sara did & not, aut, ographs, bags, outlets \\
& , cut the bread, and taste the cheese. But she did & not, Play, ooters, Net, bags \\
& and loud. They did & not, pered, cher, communities, angles \\
& fun. They did & not, orb, iour, cher, recounted \\
& & \\
\multirow{5}{*}{\textbf{64 (0.179)}} & very scared. She did not know what to & do, eat, cook, wash, pack \\
& . "Don't worry, we'll & find, go, get, fix, clean \\
& Sam," said Tim. "Do you want to & play, go, race, slide, ride \\
& her mom if they could & go, play, buy, have, make \\
& Tom's faces. "You two need to & learn, go, find, hurry, clean \\
& & \\
\multirow{5}{*}{\textbf{65 (0.243)}} & floor. They are sorry. They do & not, Wr, vanished, ch, choke \\
& might fall in!" Ben did & not, 't, generation, cled, ographs \\
& had to pick some onions for dinner. Sara did & not, wrong, unlucky, uncomfortable, uneasy \\
& ask God to help you eat your soup." Tom did & not, 't, lier, Winner, lers \\
& blue crayon and strike the wall." Ben does & not, bags, earnings, Village, lers \\
& & \\
\multirow{5}{*}{\textbf{66 (0.064)}} & Jack said, "Sure, that would be great!" The little & girl, boy, ably, ched, orers \\
& red, orange, and yellow colors. One day, a little & girl, boy, scientists, acity, antly \\
& help her whenever she needed it. And the little & girl, boy, rolled, anted, use \\
& was a little & girl, boy, ations, ators, pots \\
& was a little & girl, boy, ations, ators, pots \\
& & \\
\multirow{5}{*}{\textbf{67 (0.161)}} & found you!" It was her friend, Tim.  Lily gigg & les, ly, ling, le, showed \\
& saw that Lily was suffering because she lost & the, all, a, some, something \\
& the bird. They took the bird home and cared for & him, her, the, all, many \\
& She touched the rubber duck and felt it squeak. She thought & ,, maybe, about, for, of \\
& 't want to play with him. She ignored & her, the, them, it, his \\
& & \\
\multirow{5}{*}{\textbf{68 (0.042)}} & Timmy didn & , not, t, never, on \\
& my didn & , not, t, never, 's \\
& time, Roxy didn & , not, t, `, ´ \\
& my didn & , not, t, never, 's \\
& didn & , not, t, 's, . \\
& & \\
\multirow{5}{*}{\textbf{69 (0.084)}} & clumsy, Sam," Tom said, shaking his & hand, fist, finger, tail, arm \\
& the rain. She would jump in all the pudd & les, rejo, equal, defender, Matthew \\
& found you!" It was her friend, Tim.  Lily gigg & led, sacked, decreased, yielded, uted \\
& empty. She frown & s, ged, outs, ced, fully \\
& watch where he was going and tri & pped, led, sank, ave, annah \\
& & \\
\multirow{5}{*}{\textbf{70 (0.018)}} & his friends. One & of, was, sunny, ,, friend \\
& friends. One & of, was, sunny, ,, friend \\
& under her plate or give them to the dog. One & night, of, morning, time, sunny \\
& . One & of, was, ,, morning, is \\
& the park with her friends. One & of, was, ,, night, sunny \\
& & \\
\multirow{5}{*}{\textbf{71 (0.080)}} & angry. Lily and & Ben, Tom, Jill, Mint, Fay \\
& ," Tom said. Lily and & Tom, itt, est, hy, ippers \\
& It had a cut on its leg. Lily and & Ben, Tom, Mint, Flor, Shawn \\
& Anna and & Ben, iner, ability, astical, sub \\
& Lily and & Ben, Tom, Jack, Mark, Peter \\
& & \\
\multirow{5}{*}{\textbf{72 (0.065)}} & ?" Mom asked. Lily and & Max, Tom, Lily, her, Tim \\
& angry. Lily and & Max, her, Tom, the, Tim \\
& grandma. She misses us a lot." Lily and & her, Max, Lily, Mom, mom \\
& happy." Anna and & her, Tom, the, Max, Lily \\
& Anna and & her, Tom, Lily, the, Max \\
& & \\
\multirow{5}{*}{\textbf{73 (0.065)}} & a time, & in, a, the, they, it \\
& a time, & in, a, the, they, it \\
& a time, & in, a, the, they, it \\
& a time, & in, a, the, they, it \\
& a time, & in, a, the, they, it \\
& & \\
\multirow{5}{*}{\textbf{74 (0.153)}} & leaves under her feet and tried to climb the icy hill again. This & time, ines, ans, mong, neys \\
& that he needed to be more & comfortable, organized, ., flexible, independent \\
& on, Max made sure to watch where he was going and to be more & comfortable, flexible, obedient, independent, graceful \\
& wife and said, "I will always provide for & you, ainer, ol, Out, ooked \\
& . He loves his sister. He says, "I am sorry, Anna. & I, Will, Sorry, Hi, In \\
& & \\
\multirow{5}{*}{\textbf{75 (0.147)}} & walking towards him. He was so scared that he didn't know what to & do, see, stir, sound, step \\
& didn't know it would be so noisy." Lily forgave him and they & continued, gigg, resumed, repeated, stared \\
& very scared. She did not know what to & do, think, see, hear, smell \\
& to unravel and Timmy and Sally didn't know what to & do, think, say, see, finish \\
& for your body." Benny listened to Ollie's & story, wise, song, words, voice \\
& & \\
\multirow{5}{*}{\textbf{76 (0.411)}} & They like to play with their toys and books & in, and, ,, together, ," \\
& day, Timmy went to play with his friends in the park & ,, and, with, again, for \\
& . Max loved to play with his friends at the park & ,, every, and, because, with \\
& are friends. They like to play in the park & with, and, every, near, , \\
& had a big toy that she really wanted & to, ,, and, !, but \\
& & \\
\multirow{5}{*}{\textbf{77 (0.073)}} & Tom felt sad and angry. He wanted to make Lily share. He had an & idea, island, tale, islands, kins \\
& It's flying very far away." Max w & igg, add, ags, aded, ailed \\
& Then, Lily's daddy had an & idea, kins, ges, bows, ters \\
& it. Billy said, "I have an & idea, kins, bows, ters, leen \\
& told him about his problem.  The rabbit had an & idea, kins, bows, ers, ters \\
& & \\
\multirow{5}{*}{\textbf{78 (0.057)}} & her mommy and daddy. One day, when they went to see the ze & od, oise, ric, zag, in \\
& hill and into the pond. Timmy and his friends laughed and had so much & more, time, that, to, energy \\
& Lily's mom asked her if she wanted to have a fancy tea & set, with, ., tea, place \\
& you for the treat!" Spot bark & ed, agged, fed, apped, led \\
& in a small house, there lived a kind and & compassionate, humble, modest, poor, harmless \\
& & \\
\multirow{5}{*}{\textbf{79 (0.065)}} & corn move back and & the, it, they, down, he \\
& wash it with soap and & soap, put, a, scrub, make \\
& to play with their blocks and & dolls, their, share, books, have \\
& Lily decorated it with sweet frosting and & colorful, candles, glitter, lots, spark \\
& jump and & play, have, the, catch, see \\
& & \\
\multirow{5}{*}{\textbf{80 (0.012)}} & shoes before going outside to play.Once upon a & week, few, while, day, little \\
& that might have something yummy inside.Once upon a & week, few, day, while, long \\
& pond, happy and clean. The end.Once upon a & week, long, few, beautiful, day \\
& to his mom and be careful when playing outside.Once upon a & week, day, few, nice, little \\
& be extra careful not to bite anyone again.Once upon a & few, week, little, while, long \\
& & \\
\multirow{5}{*}{\textbf{81 (0.071)}} & she should have been more careful. From that & day, cers, acas, neys, umm \\
& too because she helped the bird. From that & day, umm, ts, per, acas \\
& up on the fridge. From that & day, ters, anes, acas, ations \\
& on her finger to make it feel better. From that & day, saf, circus, concert, lectures \\
& sharing all of their toy tools. From that & day, umm, sters, Wings, Balls \\
& & \\
\multirow{5}{*}{\textbf{82 (0.261)}} & teddy bear. It is soft and brown. It & is, likes, looks, makes, does \\
& on the swings and the & slides, swings, squirrel, other, sees \\
& every day. The plant had green & plants, roots, grass, stems, and \\
& see many things inside. There are books, toys, clothes, and & a, more, even, games, food \\
& finds a small toy car with & no, the, three, his, many \\
& & \\
\multirow{5}{*}{\textbf{83 (0.092)}} & in the future.Once upon a time, there was a big elephant named & Ellie, Mighty, George, Harry, Daisy \\
& pond. The duck sees the ball and swim & s, olds, m, ets, ases \\
& Spot ran to get it. They both laughed when Spot accidentally knocked over a be & aver, ak, ep, aker, at \\
& "I'm sorry. Will you forgive me?" Her friend thought about & this, what, the, that, how \\
& from the dangerous land.Once upon a time, there was a big dog named & Spot, Tom, Buddy, Rex, Bark \\
& & \\
\multirow{5}{*}{\textbf{84 (0.008)}} & them disappear again.Once upon a time, there was a little girl named & Lily, L, Sara, Spirit, Inf \\
& upon a time, there was a little girl named & Lily, L, Sara, D, Sandy \\
& Once upon a time, there was a little girl named & Lily, L, Sara, Daisy, Anna \\
& ever frightened again.Once upon a time, there was a little girl named & Lily, L, Sara, Po, D \\
& Once upon a time, there was a little boy named & Tom, Tommy, Ben, Sam, Bob \\
& & \\
\multirow{5}{*}{\textbf{85 (0.073)}} & Anna and Ben are playing with cray & ons, hers, iers, od, eter \\
& The spider was angry and chased after Buzz. Buzz crawled as fast as he & could, boy, girl, E, cer \\
& to unravel and Timmy and Sally didn't know what to & say, expect, think, did, use \\
& acorn. The moral of the & day, lesson, joke, game, lessons \\
& up and continued to play games together, but this time, Max made & a, the, it, up, his \\
& & \\
\multirow{5}{*}{\textbf{86 (0.110)}} & proud of herself for helping her furry friend.Once upon a time & there, at, in, later, it \\
& listen to her mom and always be safe.Once upon a time & there, in, at, it, they \\
& under her plate or give them to the dog. One day & she, the, when, they, her \\
& friends. They played together every day. One day & the, it, they, Tim, Tom \\
& importance of sharing and being kind to his friends.Once upon a time & there, at, in, later, with \\
& & \\
\multirow{5}{*}{\textbf{87 (0.126)}} & play with her friends. & One, They, She, Yesterday, Do \\
& her mommy and daddy. & One, They, Yesterday, Do, Grace \\
& , Anna was feeling bossy. & She, First, Lisa, Jenna, Mark \\
& to her bed. & She, It, One, When, Every \\
& play together in the big green park near their house. & One, They, There, Sally, Tommy \\
& & \\
\multirow{5}{*}{\textbf{88 (0.113)}} & his ball into the goal. Spot ran fast with the ball in & the, its, one, her, front \\
& noise. It was a car that zoom & past, ing, by, !, across \\
& noises. Tom had a small car that could go fast and be & loud, fast, very, slow, eps \\
& said. "Deal, Mom. Thank you, Mom. You're & welcome, very, right, a, good \\
& , red ball in the park. He threw it up high and caught it with & a, the, ease, two, one \\
& & \\
\multirow{5}{*}{\textbf{89 (0.190)}} & his ball. He walked and & talked, played, ran, looked, jumped \\
& Max saw a big plane flying in the sky. Max bark & ed, ked, ingly, fully, de \\
& She sees the letter. It is torn. She sigh & s, ers, aks, ses, rs \\
& "It's okay, my & loves, sweet, loved, buddy, just \\
& her if she shared her cereal with Timmy. Lily said yes, & but, excited, after, saying, offering \\
& & \\
\multirow{5}{*}{\textbf{90 (0.049)}} & dog running after the car.  L & ily, Lily, Linda, Lena, Rose \\
& Emma heard her sister's scream and asked, "L & ily, Lily, Ben, Anna, recovery \\
& lit up with bright lights.  L & ily, L, Linda, Lily, Rose \\
& said, "L & ily, Lily, Ben, Lena, pollen \\
& "Be careful, the edges are sharp!"  L & ily, Lily, Liam, Ben, Rose \\
& & \\
\multirow{5}{*}{\textbf{91 (0.076)}} & found you!" It was her friend, Tim.  Lily gigg & led, oured, ingly, iot, connectors \\
& lots of fun pudd & les, as, ocks, led, is \\
& to investigate and found shiny rocks that spark & led, edly, ez, lling, rying \\
& so pretty and spark & ly, Bench, Giants, RO, Sav \\
& sprink & les, led, angles, Cam, Crit \\
& & \\
\multirow{5}{*}{\textbf{92 (0.196)}} & it, but it was too heavy. The barrel rolled all the & way, forest, place, jungle, mountains \\
& and see the world outside the & gate, world, garden, city, forest \\
& vanished! Timmy looked all around his & house, garden, backyard, yard, town \\
& my foot hurts. The frame fell on it." Em & ma, enny, lee, am, erson \\
& is better than fighting. And they all became good & friends, ls, behold, uld, Int \\
& & \\
\multirow{5}{*}{\textbf{93 (0.158)}} & to play in the water. He would jump and splash in the big p & uddle, water, ashes, ail, waves \\
& went to the park with her mom and saw her & little, new, favorite, daughter, toy \\
& Ben. He is sad and bored. He misses Ben a & long, chance, day, time, fun \\
& the garden. They liked to observe the bugs and the & flowers, plants, trees, bugs, worms \\
& When they got home, Lily put on her purple p & anda, endant, uddle, ears, ail \\
& & \\
\multirow{5}{*}{\textbf{94 (0.220)}} & and said, "Yes, I can help you. But first, we have & a, something, some, enough, no \\
& but Rex blocked his way. "Leave me alone! I just & wanted, wants, like, need, moved \\
& , you can," Lily and Tom said, nodding. "But you have & been, a, something, no, too \\
& asked Fluffy. "Yes, I want to come with & us, me, the, your, my \\
& said to the plant, "We are sorry, plant. We did & a, something, it, our, wrong \\
& & \\
\multirow{5}{*}{\textbf{95 (0.075)}} & in the park.Once upon a time, there was a little girl & named, lived, aked, Camer, topics \\
& 'll like them at first.Once upon a time, there was a little girl & named, Camer, irs, unks, orns \\
& can always try again tomorrow.Once upon a time, there was a little girl & named, Camer, topics, unks, ures \\
& on stage too.Once upon a time, there was a little boy & named, orns, osity, topics, unks \\
& up when things get hard.Once upon a time, there was a little girl & named, Camer, unks, irs, Prepar \\
& & \\
\multirow{5}{*}{\textbf{96 (0.066)}} & Then, Lily's daddy had an & idea, ising, ID, ep, chairs \\
& The dog stopped being frightened and started w & agging, agg, ashing, ogging, inking \\
& Tom felt sad and angry. He wanted to make Lily share. He had an & idea, example, adjective, ID, error \\
& didn't know it would be so noisy." Lily forg & ave, apped, aked, aws, understood \\
& . Sam wanted to help his friend feel better. Sam had an & idea, information, ID, ising, adjective \\
& & \\
\multirow{5}{*}{\textbf{97 (0.102)}} & had their wand and their bubbles. They did & not, ales, aper, ann, nce \\
& sorry, Mia. I wanted to win. I did & not, ator, interacted, rig, alks \\
& named Lily. She loved to play with her dolls, & but, especially, even, which, so \\
& fun. They did & not, orb, aper, iour, nce \\
& liked her tank very much and did & not, ales, uffed, plet, uned \\
& & \\
\multirow{5}{*}{\textbf{98 (0.053)}} & "You see, Mitt & ens, uffy, uff, bles, ruff \\
& The swan nodded and sw & an, ans, uttered, atted, acked \\
& get his ac & orns, rob, robat, anuts, ockey \\
& asked Fl & uffy, oppy, uff, utter, opsy \\
& I'm the last of my family. The other sw & ans, amps, anes, ippers, ooters \\
& & \\
\multirow{5}{*}{\textbf{99 (0.172)}} & Anna and Ben are playing with cray & on, ins, s, ries, els \\
& the broken jar and the crumbs on the & floor, table, ground, kitchen, sidewalk \\
& went to the circus again.Once upon a time, there & was, named, iced, class, watching \\
& grass. They were very happy. But on the & way, day, weekend, morning, evening \\
& them broke and spilled on the & floor, kitchen, stairs, sidewalk, street \\
\end{longtable}
\normalsize

\subsubsection{Mobilenet-v2-small Full Decomposition}\label{fig:cnn_full_circuits}
The most affected tokens (final token in each text) for each of the subnetworks in the mobilenet-v3-small decomposition, and the most affected logits of each sample.
\includepdf[pages=-, fitpaper=true,width=\textwidth]{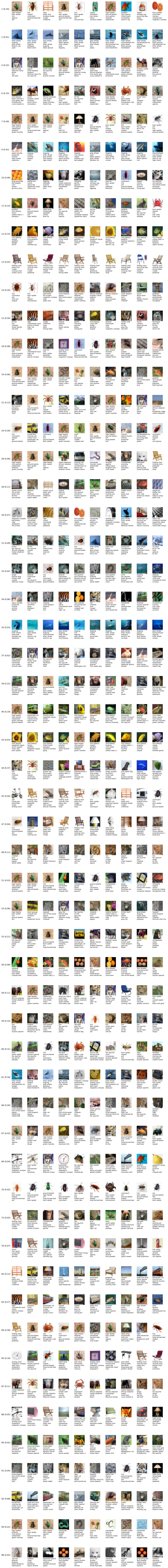}\label{fig:cnn_all}
\end{document}